\documentclass{article}

\PassOptionsToPackage{table}{xcolor}

\usepackage{microtype}
 \usepackage{bbm} 
\usepackage{graphicx}
\usepackage{multirow}
\usepackage{subfigure}
\usepackage{booktabs} 
\usepackage{longtable} 
\usepackage{hyperref}

\usepackage{tcolorbox}
\tcbuselibrary{skins, breakable}
\usepackage[utf8]{inputenc}  
\usepackage[T1]{fontenc}   
\usepackage{textcomp}      
\usepackage{xcolor}
\usepackage{listings}
\usepackage[accepted]{icml2025}

\usepackage{amsmath}
\usepackage{amssymb}
\usepackage{mathtools}
\usepackage{amsthm}

\usepackage[capitalize,noabbrev]{cleveref}

\theoremstyle{plain}

\theoremstyle{definition}

\theoremstyle{remark}

\usepackage[textsize=tiny]{todonotes}

\icmltitlerunning{MedProbeBench}

\begin{document}

\twocolumn[
\icmltitle{MedProbeBench: Systematic Benchmarking at \\
           Deep Evidence Integration for Expert-level Medical Guideline}

\icmlsetsymbol{equal}{*}
\icmlsetsymbol{corr}{\ensuremath{\dagger}}

\begin{icmlauthorlist}
\icmlauthor{Jiyao Liu}{fdu,sail,equal}
\icmlauthor{Jianghan Shen}{nju,sail,equal}
\icmlauthor{Sida Song}{fdu,equal}
\icmlauthor{Tianbin Li}{sail}
\icmlauthor{Xiaojia Liu}{fdu}
\icmlauthor{Rongbin Li}{sail}
\icmlauthor{Ziyan Huang}{sail}
\icmlauthor{Jiashi Lin}{sail}
\icmlauthor{Junzhi Ning}{sail}
\icmlauthor{Changkai Ji}{fdu,sail}
\icmlauthor{Siqi Luo}{sail}
\icmlauthor{Wenjie Li}{sail} 
\icmlauthor{Chenglong Ma}{fdu,sail}
\icmlauthor{Ming Hu}{sail}
\icmlauthor{Jing Xiong}{hku}
\icmlauthor{Jin Ye}{sail}
\icmlauthor{Bin Fu}{sail}
\icmlauthor{Ningsheng Xu}{fdu}
\icmlauthor{Yirong Chen}{sail}
\icmlauthor{Lei Jin}{fdu}
\icmlauthor{Hong Chen}{fdu,corr}
\icmlauthor{Junjun He}{sail,corr}
\end{icmlauthorlist}

\icmlaffiliation{fdu}{Fudan University, Shanghai, China}
\icmlaffiliation{nju}{Nanjing University, Nanjing, China}
\icmlaffiliation{sail}{Shanghai Artificial Intelligence Laboratory, Shanghai, China}
\icmlaffiliation{hku}{University of Hong Kong}

\icmlcorrespondingauthor{Hong Chen}{chenhong@huashan.org.cn}
\icmlcorrespondingauthor{Junjun He}{hejunjun@pjlab.org.cn}

\icmlkeywords{Large Language Models, Medical AI, Benchmarking, Deep Research, Evidence Integration, Citation Verification}

\vskip 0.3in
]

\printAffiliationsAndNotice{\icmlEqualContribution}  

\begin{abstract}
Recent advances in deep research systems enable large language models to retrieve, synthesize, and reason over large-scale external knowledge. In medicine, developing clinical guidelines critically depends on such deep evidence integration. However, existing benchmarks fail to evaluate this capability in realistic workflows requiring multi-step evidence integration and expert-level judgment.
To address this gap, we introduce \textbf{MedProbeBench}, the first benchmark leveraging high-quality clinical guidelines as expert-level references. Medical guidelines, with their rigorous standards in neutrality and verifiability, represent the pinnacle of medical expertise and pose substantial challenges for deep research agents.
For evaluation, we propose \textbf{MedProbe-Eval}, a comprehensive evaluation framework featuring: \textit{(1) Holistic Rubrics} with 1,200+ task-adaptive rubric criteria for comprehensive quality assessment and \textit{(2) Fine-grained Evidence Verification} for rigorous validation of evidence precision, grounded in 5,130+ atomic claims. Evaluation of 17 LLMs and deep research agents reveals critical gaps in evidence integration and guideline generation, underscoring the substantial distance between current capabilities and expert-level clinical guideline development. Project \& Dataset: \href{https://github.com/uni-medical/MedProbeBench}{https://github.com/uni-medical/MedProbeBench}
\end{abstract}

\section{Introduction}
\label{sec:intro}

\begin{figure}[!htb]
\centering
\vspace{5mm}
\includegraphics[width=1\columnwidth]{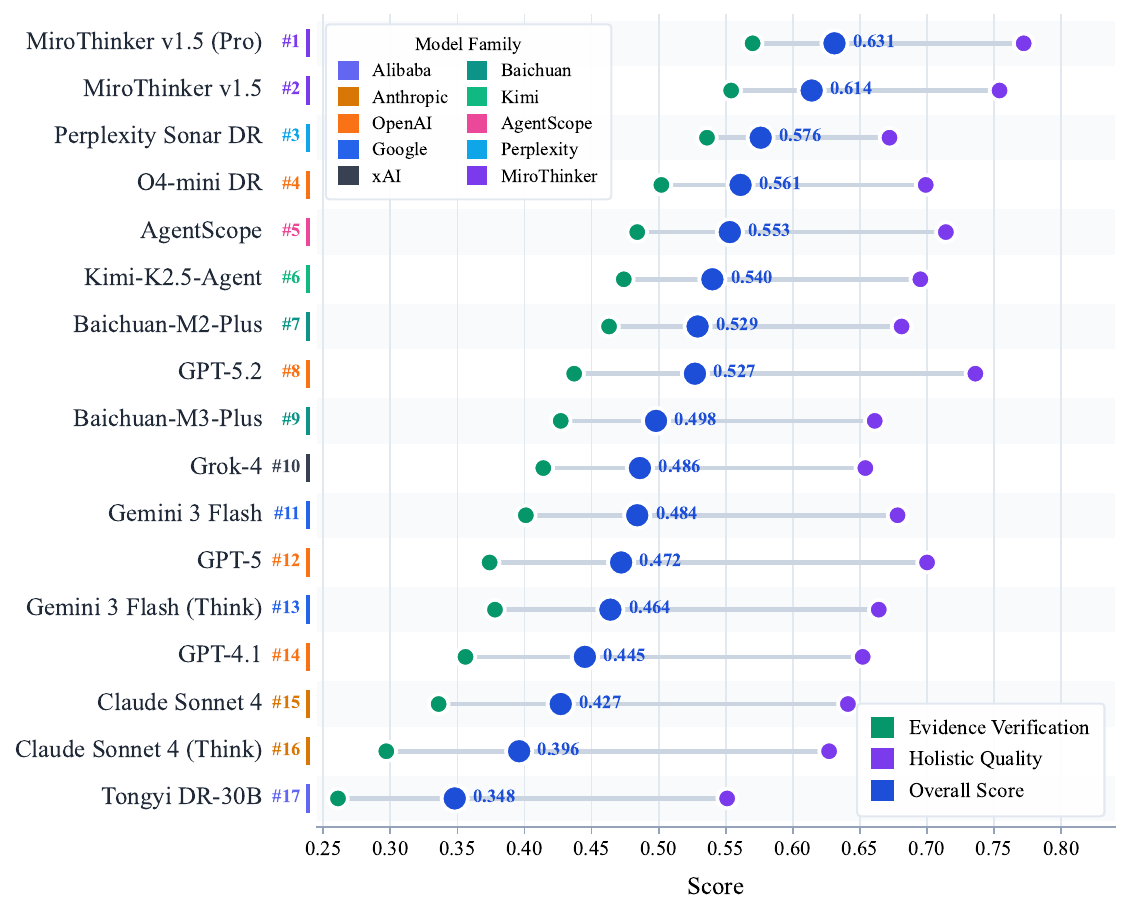}
\vspace{-3mm}
\caption{Performance of 17 LLMs and deep research agents on MedProbeBench. Each model is evaluated on holistic quality and fine-grained evidence verification. Even the best-performing deep research agent achieves only 0.631 overall, revealing a significant gap toward expert-level clinical guideline generation.}
\vspace{-1mm}
\label{fig:overall}
\end{figure} 

AI agents are increasingly deployed to tackle \emph{expert-level medical tasks} that demand specialized domain knowledge and multi-step reasoning, including clinical decision support~\cite{yang2025medaide}, diagnostic reasoning~\cite{wang2025medagent,su2025gmai,xu2025medground}, and treatment-oriented medical inference workflows. Recent systems demonstrate the potential of general-purpose biomedical AI agents across these diverse tasks~\cite{huang2025biomni}. Among these, \emph{deep research agents}~\cite{du2025deepresearch} represent one of the most advanced agent architectures, capable of autonomously planning, searching, and synthesizing information to complete research tasks that would typically require significant time and effort from human experts.

\begin{figure*}[!ht]
  \centering
  \includegraphics[width=\textwidth]{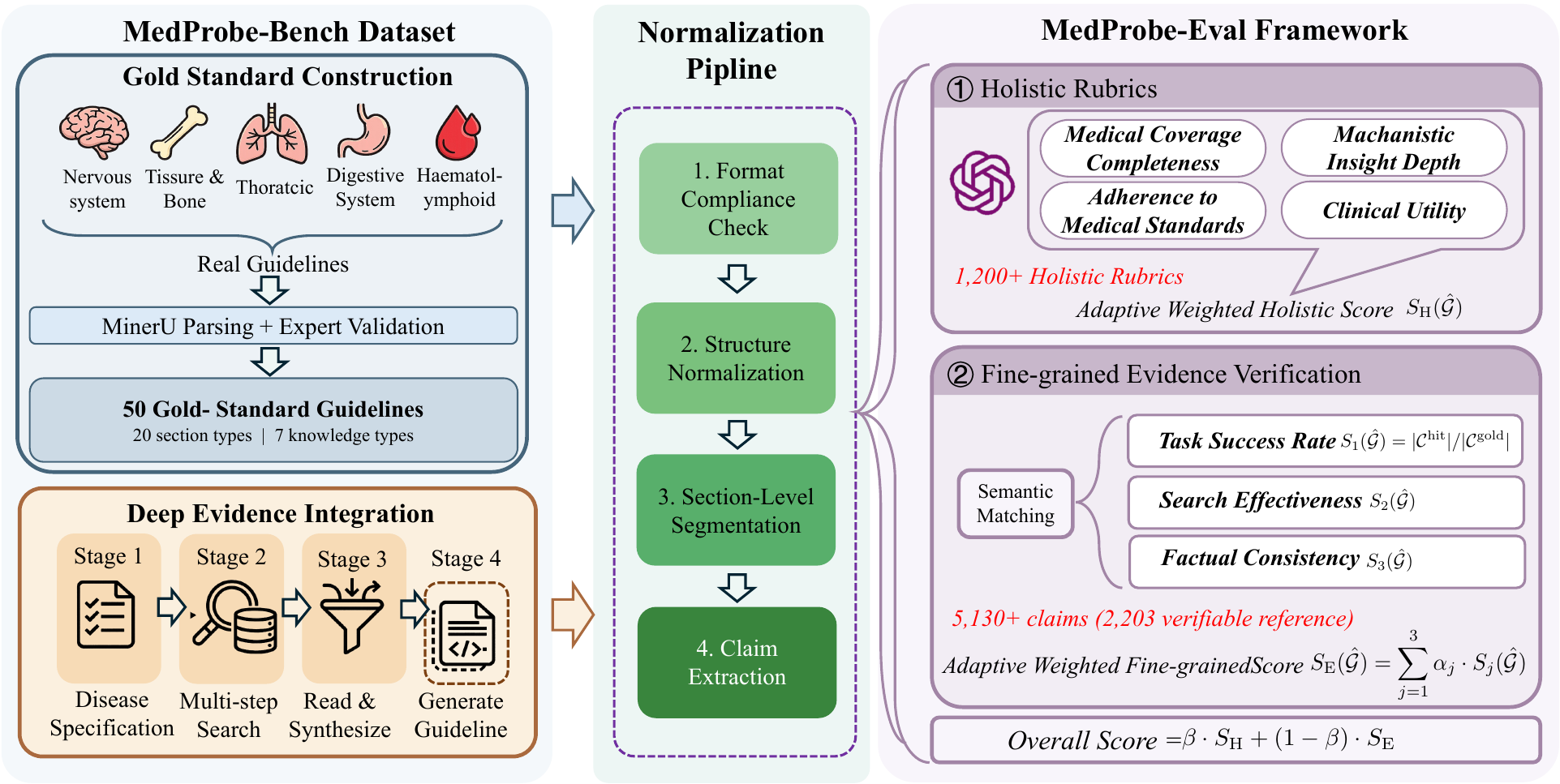}
  \vspace{-3mm}
  \caption{Ground-truth clinical guidelines and model-generated reports are processed through an identical normalization and structuring pipeline and subsequently evaluated using a two-stage framework comprising holistic rubric-based quality assessment and fine-grained evidence verification. Together. MedProbeBench enables systematic analysis of long-horizon deep evidence integration.}
  \label{fig:method_pipeline}
\end{figure*}

Recent research has made initial attempts to evaluate medical deep evidence integration capabilities through various agent benchmarks. For instance, MedBrowseComp~\cite{chen2025medbrowsecomp} tests multi-hop fact retrieval from medical knowledge bases, and MIRAGE~\cite{xiong2024benchmarking} assesses retrieval-augmented generation performance on medical QA datasets.
However, these benchmarks remain insufficient for evaluating expert-level medical synthesis capabilities due to several critical limitations:
\textit{(1) Limited evaluation of complex expert report generation.} Existing benchmarks primarily focus on short-form question answering or isolated clinical reasoning tasks, failing to assess the ability to generate comprehensive, multi-section expert documents that mirror real-world guideline development workflows.
\textit{(2) Simplistic evaluation formats.} Current benchmarks like MedBrowseComp~\cite{chen2025medbrowsecomp} evaluates through multiple-choice or short-answer formats, which cannot capture the complexity of long-horizon evidence synthesis and structured medical writing.
\textit{(3) Lack of expert-level quality assessment.} Most benchmarks lack systematic evaluation frameworks grounded in clinical standards (e.g., medical taxonomy adherence, clinical terminology precision), making it difficult to assess whether generated content meets professional guideline requirements.
\textit{(4) Absence of fine-grained evidence verification.} Existing evaluations primarily assess outputs holistically without claim-level evidence verification, making it difficult to distinguish genuine evidence synthesis from parametric memorization or hallucinated reasoning.

To address these limitations, we introduce \textbf{MedProbeBench}, the first benchmark that leverages high-quality clinical guidelines as expert-level references to evaluate deep evidence integration capabilities. Clinical guidelines, with their rigorous standards in neutrality, professionalism, and verifiability, represent the pinnacle of medical expertise and pose substantial challenges for deep research agents.
Concretely, MedProbeBench comprises 50 physician-validated clinical guidelines spanning 5 medical specialties, with 5,133 atomic claims (of which 2,203 carry verifiable reference anchors). For evaluation, we propose \textbf{MedProbe-Eval}, a dual-tier assessment framework featuring: \textit{(1) Holistic Rubrics} with 1,236 task-adaptive criteria across four clinically grounded dimensions to assess overall guideline quality, and \textit{(2) Fine-grained Evidence Verification} with three-dimensional reliability metrics to verify claim-level evidence grounding.
As shown in Figure~\ref{fig:overall}, our evaluation of 17 state-of-the-art systems reveals that even the best-performing agent achieves only 0.631 overall, highlighting a significant gap toward expert-level guideline generation. The contributions of this work are threefold:
\begin{enumerate}
\item \textbf{MedProbe-Dataset.} A curated dataset of physician-validated clinical guidelines with claim-level annotations, designed to evaluate deep evidence integration in expert-level medical synthesis.
\item \textbf{MedProbe-Eval.} A dual-tier evaluation framework combining holistic rubric-based quality assessment with fine-grained evidence verification, enabling systematic and reproducible benchmarking of medical deep research agents.
\item \textbf{Comprehensive evaluation and benchmark maintenance.} Extensive experiments across 17 diverse systems with thorough analysis and human studies to validate our framework's reliability, along with a commitment to maintaining and expanding this benchmark to reflect evolving real-world conditions.
\end{enumerate}

\section{Related Work}
\label{sec:related}

\subsection{Medical LLM Benchmarks}

Early medical LLM benchmarks predominantly evaluate closed-book or short-form question answering, exemplified by MedQA~\cite{jin2021disease}, MMLU-Medical~\cite{hendrycks2021measuring}, MedQ-Bench~\cite{liu2025medq} and PubMedQA~\cite{jin2019pubmedqa}.
While effective for assessing factual knowledge, these benchmarks emphasize isolated recall and have largely reached performance saturation on frontier models.
A recent systematic review~\cite{medbenchreview2025} identifies a core limitation: most evaluations assess knowledge in isolation rather than practice-oriented workflows that reflect real clinical expertise.
Subsequent benchmarks introduce greater task complexity or retrieval augmentation~\cite{zuo2025medxpertqa,arora2025healthbench,ning2026mmrarebench,xiong2024benchmarking}, yet continue to focus on answering questions with definitive answers.
Several recent efforts move toward guideline-aware or evidence-grounded evaluation. AMEGA~\cite{fast2024autonomous} evaluates LLM adherence to medical guidelines through open-ended clinical scenarios across 13 specialties, but assesses compliance with existing guidelines rather than the ability to synthesize and generate them.
MedGUIDE~\cite{li2025medguide} benchmarks clinical decision-making against 55 NCCN oncology decision trees using multiple-choice questions, focusing on protocol adherence rather than open-ended document generation.
MedR-Bench~\cite{qiu2025quantifying} assesses evidence-based multi-step reasoning across 13 body systems using structured patient cases, targeting diagnostic reasoning quality rather than long-form guideline synthesis with claim-level citation verification.
In contrast, MedProbeBench evaluates deep evidence integration through structured, long-horizon clinical guideline generation requiring multi-source synthesis with explicit, claim-level citation accountability.

\subsection{Deep Research Systems and Benchmarks}
Deep research systems support multi-step information gathering and synthesis beyond single-turn question answering.
Early work such as WebGPT~\cite{nakano2021webgpt}, together with subsequent commercial systems including Perplexity Deep Research~\cite{perplexity2025deepresearch}, OpenAI Deep Research~\cite{openai2024deepresearch}, and Gemini Deep Research~\cite{google2024geminiresearch}, demonstrates the effectiveness of iterative retrieval and long-form synthesis.
Agent-based medical systems such as AURA~\cite{aura2025}, MedAgent-Pro~\cite{medagentpro2025}, and TxAgent~\cite{txagent2025} further demonstrate advanced multi-step reasoning and tool use in clinical domains, while Tongyi DeepResearch~\cite{team2025tongyi} and MiroThinker~\cite{team2025mirothinker} extend these capabilities through longer contexts and interactive reasoning.
Benchmarks such as ResearchRubrics~\cite{sharma2025researchrubrics}, DeepResearch Bench~\cite{du2025deepresearch}, and BrowseComp~\cite{wei2025browsecomp} aim to standardize the evaluation of deep research agents, but remain largely domain-agnostic and lack medical-specific evaluation criteria.
MedProbeBench bridges this gap by targeting medical deep research agents with structured, long-horizon clinical guideline generation and claim-level citation verification.

\section{Constructing MedProbeBench}
\label{sec:method}

\subsection{Task Definition: Deep Evidence Integration}

MedProbeBench evaluates long-form deep evidence integration, an expert-level medical synthesis task that requires AI systems to construct a comprehensive, evidence-grounded clinical guideline through systematic retrieval, analysis, synthesis, and citation of authoritative sources.

Formally, given a disease specification $\mathcal{D}$, accessible knowledge sources $\mathcal{K}$, and a structured schema $\mathcal{S} = \{s_1, \ldots, s_n\}$ defining required content sections, the system must generate 
a document $\hat{\mathcal{G}}$ that satisfies three core criteria: (i)~\emph{structural completeness}, requiring adherence to the predefined guideline schema and coverage of all clinically required sections; (ii)~\emph{terminological precision}, requiring consistency with authoritative classification systems and established medical conventions; and (iii)~\emph{evidence grounding}, requiring that each key claim is supported by verifiable references. Here, key claims encompass all atomic clinical statements of recognized types, including Factual, Mechanistic, Clinical, Diagnostic, Differential, Prognostic, and Therapeutic claims.

\subsection{Gold Standard Collection}
To establish reliable ground truth for evaluation, we curate gold-standard documents from the WHO Classification of Tumours (5th Edition), one of the most authoritative and globally adopted clinical reference standards in oncology. We select five representative volumes covering major organ systems: (1) Central Nervous System Tumours~\cite{louis2021cimpact}; (2) Soft Tissue and Bone Tumours~\cite{who2020softtissue}; (3) Digestive System Tumours~\cite{who2019digestive}; (4) Thoracic Tumours~\cite{who2021thoracic}; (5) Haematolymphoid Tumours~\cite{who2024haematolymphoid}. These sources collectively span diverse tumor types, molecular mechanisms, and diagnostic criteria, providing multi-level complexity suitable for benchmarking long-horizon, evidence-grounded synthesis.

\subsection{Shared Report Normalization and Structuring}
To enable fair and fine-grained evaluation of long-form clinical guideline generation, we normalize both ground-truth guidelines and model-generated reports into a shared section- and claim-level representation. All documents are processed through an identical normalization and claim extraction pipeline, as shown in Figure~\ref{fig:method_pipeline}, ensuring comparison within a unified structural space and minimizing evaluation bias induced by formatting and stylistic differences.

Ground-truth guideline sources additionally undergo minimal document conversion and expert validation to ensure clinical reliability prior to normalization. Specifically, PDF and HTML guidelines are first converted to structured Markdown using the parsing tool, MinerU~\cite{mineru2024}, and reviewed by board-certified medical professionals.

Subsequently, both ground-truth and model-generated documents are processed through five core stages:
(1) Format Compliance Checking, which verifies adherence to a predefined guideline section schema;
(2) Markdown Structure Normalization, which rewrites documents into a unified Markdown format with standardized section titles and hierarchy;
(3) Section-Level Segmentation, which decomposes documents into individual sections;
(4) Guideline Claim Extraction, which extracts atomic clinical claims with associated in-text citations; and
(5) Medical Guideline Citation Normalization and Enrichment, which resolves non-canonical citations into canonical PubMed- or DOI-based references when possible.

By enforcing an identical processing pipeline and isolating ground-truth-specific interventions, we minimize structural and formatting biases in downstream evaluation and enable reliable, claim-level comparison between curated clinical guidelines and generated reports.

\subsection{Dataset Overview}
\begin{figure}[t]
  \centering
  \includegraphics[width=\columnwidth]{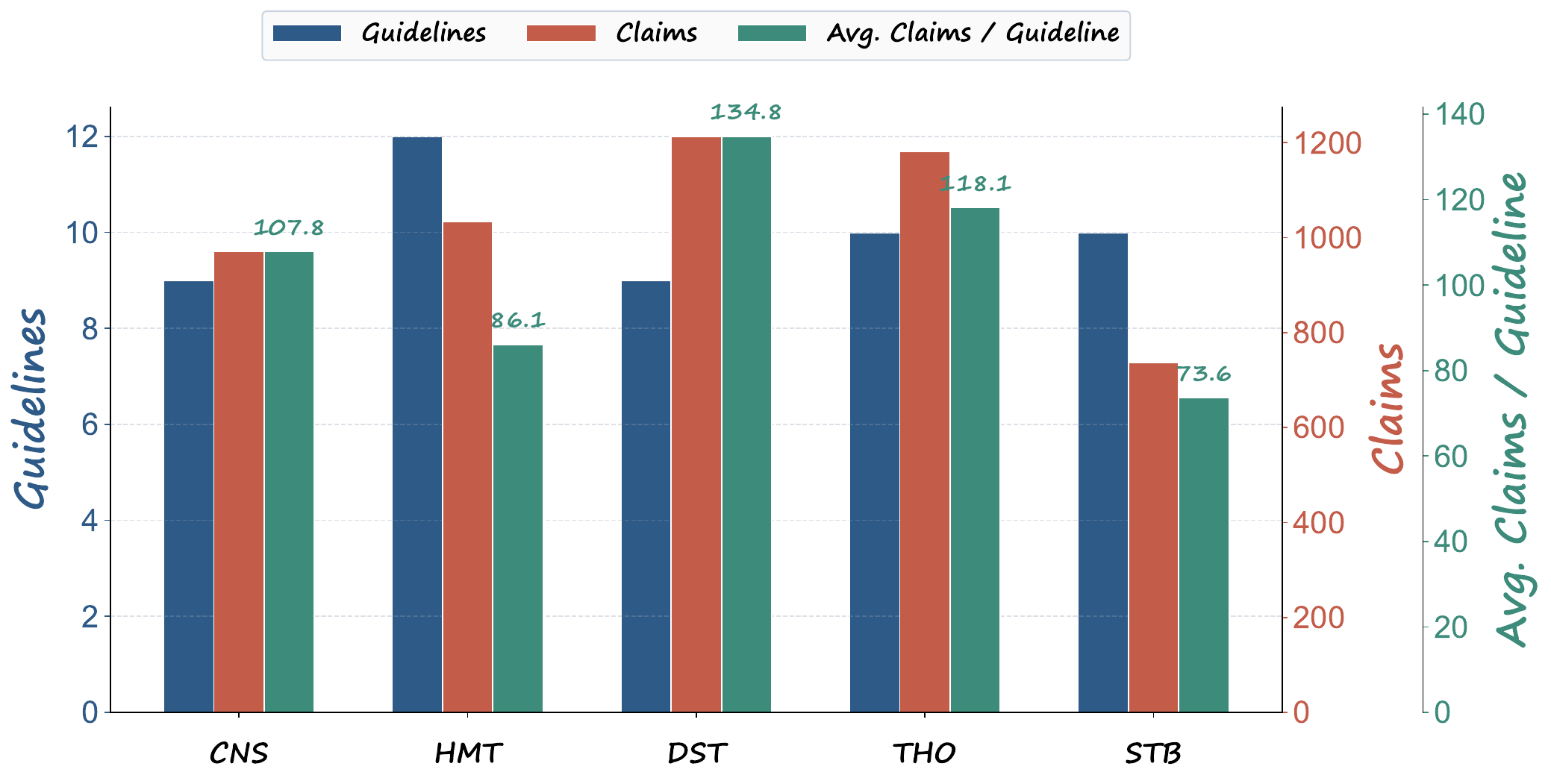}
  \caption{Dataset overview of MedProbeBench across medical domains. CNS: Central Nervous System Tumours; STB: Soft Tissue and Bone Tumours; DST: Digestive System Tumours; THO: Thoracic Tumours; HMT: Haematolymphoid Tumours.}
  \vspace{-6mm}
  \label{fig:Overview}
\end{figure}

Figure~\ref{fig:Overview} summarizes the distribution of guidelines and atomic claims across medical domains in MedProbeBench.
The benchmark spans five disease domains with a comparable number of guideline documents per domain, while exhibiting substantial variation in total claim counts and average claims per guideline, reflecting differences in guideline length and structural complexity across specialties.

Across all tasks, MedProbeBench covers 20 guideline section types, including Definition, ICD Coding, Essential Diagnostic Criteria, Localization, Macroscopic Appearance, Clinical Features, Grading / Staging, Spread, Diagnostic Molecular Pathology, Pathogenesis, Epidemiology, Prognosis and Prediction, Imaging, Immunophenotype, Histopathology, Etiology, Cytology, Subtype(s), Differential Diagnosis, and Related Terminology.
At a higher level, task content spans seven knowledge types, which detailed in Section~\ref{sec:rubric}.
Of the 5,133 total atomic claims in the benchmark, \textbf{2,203 (42.92\%)} carry at least one supporting citation, providing verifiable reference anchors for fine-grained evidence verification. Detailed benchmark-scale statistics, including holistic rubric counts and citation coverage by domain, are reported in Appendix~\ref{app:benchmark_scale}.

\section{Evaluation Framework}
\label{sec:rubric}

Our evaluation framework focuses on two key aspects: (1) \textit{holistic rubrics} for assessing overall guideline quality, and (2) \textit{fine-grained evidence verification} for evaluating retrieval accuracy and claim-level grounding. Together, these dimensions capture both expert-level synthesis quality and evidence fidelity in medical deep evidence integration tasks. Details of the LLM-based evaluation prompts used in both components are provided in Appendix~\ref{app:prompt_template}.

\subsection{Holistic Rubrics}

Given a generated guideline $\hat{\mathcal{G}}$, we evaluate its overall quality across four dimensions, denoted by $s_1(\hat{\mathcal{G}})$ through $s_4(\hat{\mathcal{G}})$, respectively:
\textit{Medical Coverage Completeness} evaluates whether the guideline comprehensively addresses key clinical aspects, including symptom profiles, diagnostic criteria, treatment modalities, and prognostic considerations.
\textit{Mechanistic Insight Depth} assesses whether the content provides mechanistic understanding at the molecular or pathological level, rather than surface-level descriptions.
\textit{Adherence to Medical Standards} measures consistency with authoritative classification systems, ensuring that terminology and categorization align with established medical consensus.
\textit{Clinical Utility} assesses whether the guidance offers actionable information that can support clinical decision-making in real-world practice settings.

The overall holistic score is computed as a weighted average:
\begin{equation}
S_{\mathrm{H}}(\hat{\mathcal{G}}) = \sum_{i=1}^{4} w_i \cdot s_i(\hat{\mathcal{G}}), \quad \text{where} \quad \sum_{i=1}^{4} w_i = 1
\end{equation}
where each dimension score $s_i(\hat{\mathcal{G}})$ is provided by an LLM-based evaluator (GPT-4.1, temperature=0) on a 0-10 scale. Following DeepResearch Bench~\citep{du2025deepresearch}, the dimension weights $w_i$ are task-adaptive: they are derived from the task specification and section schema, averaged over $T$ independent trials, and normalized to sum to one. The resulting weight vector is fixed per task and shared across all evaluated systems (details in Appendix~\ref{app:benchmark_scale}).

\subsection{Fine-grained Evidence Verification}

\textbf{Claim Decomposition.}
To enable granular evaluation of evidence grounding, we decompose each guideline into atomic, verifiable clinical claims.
Formally, we define a decomposition operator $\mathcal{D}: \mathcal{G} \rightarrow \mathcal{C}$ that maps a guideline $\mathcal{G}$ to a set of claims $\mathcal{C} = \{c_1, c_2, \ldots, c_n\}$.
Each claim is represented as a triple $c_i = (t_i, \mathcal{R}_i, \tau_i)$, where $t_i$ is the claim text, $\mathcal{R}_i$ is the set of supporting references cited for that claim, and $\tau_i$ is the claim type drawn from seven clinically meaningful categories: Factual, Mechanistic, Clinical, Diagnostic, Differential, Prognostic, and Therapeutic.
In practice, $\mathcal{D}$ is implemented as an LLM-based extractor: given each normalized section of $\mathcal{G}$, a language model identifies all verifiable atomic statements, associates each with its in-text citation markers, and assigns a claim type. Compound sentences are split into minimal, independently verifiable units, while quantitative values, gene aliases, and domain-specific annotations are preserved verbatim. The full extraction prompt is provided in Appendix~\ref{app:claim_extraction}.

\textbf{Semantic Matching.}
We define a semantic matching relation $\sim$ between individual claims: $c_1 \sim c_2$ if and only if $\mathrm{sim}(t_1, t_2) \ge \theta$, where $t_i$ denotes the textual component of claim $c_i$, and $\mathrm{sim}(\cdot,\cdot)\!:[0,1]$ is a semantic similarity function over claim texts.
A generated claim $c^{\mathrm{gen}}$ is then said to be \emph{matched against} a gold claim set $\mathcal{C}^{\mathrm{gold}}$ if
\begin{equation}
\exists\, c^{\mathrm{gold}} \in \mathcal{C}^{\mathrm{gold}} : c^{\mathrm{gen}} \sim c^{\mathrm{gold}},
\end{equation}
i.e., there exists at least one gold claim individually matched by $c^{\mathrm{gen}}$.
We use $c^{\mathrm{gen}} \sim \mathcal{C}^{\mathrm{gold}}$ as shorthand for this existential condition; note that this is a relation between a claim and a set, not an equivalence relation.
where $\mathrm{sim}(\cdot,\cdot)\!:\mathcal{T}\times\mathcal{T}\rightarrow[0,1]$ denotes a semantic similarity function.

Similarity is computed using a two-stage strategy, designed to balance computational efficiency and semantic accuracy. 
We first apply a fast Jaccard similarity between the two claim texts as a pre-filter: claims with high lexical overlap are matched directly, while all remaining cases are passed to an LLM-based semantic evaluator to determine whether the generated claim covers the core medical content of the gold claim (threshold details in Appendix~\ref{app:claim_matching}).

The set of successfully matched generated claims is defined as
\begin{equation}
\mathcal{C}^{\mathrm{hit}}
=
\left\{
c \in \mathcal{C}^{\mathrm{gen}}
\,\middle|\,
c \sim \mathcal{C}^{\mathrm{gold}}
\right\}.
\end{equation}

In practice, we apply claim-level deduplication prior to computing $\mathcal{C}^{\mathrm{hit}}$: semantically equivalent generated claims are merged into a single representative, ensuring that each gold-standard claim can be credited at most once and preventing inflated coverage from repetitive outputs.

\textbf{Evaluation Metrics.}
Given the gold-standard claim set $\mathcal{C}^{\text{gold}} = \mathcal{D}(\mathcal{G}^{\text{gold}})$ and generated claim set $\mathcal{C}^{\text{gen}} = \mathcal{D}(\hat{\mathcal{G}})$, we define three metrics:

\noindent\textbf{(1) Task Success Rate.} Measures the proportion of gold-standard claims successfully generated:
\begin{equation}
S_1(\hat{\mathcal{G}}) = \frac{|\mathcal{C}^{\text{hit}}|}{|\mathcal{C}^{\text{gold}}|}
\end{equation}
This metric reflects the system's ability to capture essential medical knowledge required by the guideline development task.

\noindent\textbf{(2) Search Effectiveness.}
This metric evaluates whether the system retrieves reference sources that correspond to those used in the gold-standard guideline. 
Instead of operating at the claim level, we aggregate references at the section level and measure reference recall based on semantic matching.

For each section $s$, let $\mathcal{R}^{\text{gold}}_s$ and $\mathcal{R}^{\text{gen}}_s$ denote the sets of reference sources appearing in the gold and generated guidelines, respectively. 
We perform one-to-one matching between references based on semantic equivalence of their underlying content (e.g., same study, dataset, or object), where each generated reference can be matched to at most one gold reference.

Let $M_s$ denote the set of successfully matched gold references in section $s$. The overall reference recall is computed as:
\begin{equation}
\text{Recall} = \frac{\sum_s |M_s|}{\sum_s |\mathcal{R}^{\text{gold}}_s|}.
\end{equation}

To account for the quantity of retrieved references, we further define a quantity score:
\begin{equation}
\text{Quantity} = \min\left(1, \frac{\sum_s |\mathcal{R}^{\text{gen}}_s|}{\sum_s |\mathcal{R}^{\text{gold}}_s|}\right).
\end{equation}

The final Search Effectiveness score is computed as a weighted combination:
\begin{equation}
S_2(\hat{\mathcal{G}}) = 0.6 \cdot \text{Recall} + 0.4 \cdot \text{Quantity}.
\end{equation}

Reference matching is performed using content-based semantic comparison rather than exact string matching. Two references are considered a match if they refer to the same underlying topic, study, dataset, or object, even if their surface forms differ.

\noindent\textbf{(3) Factual Consistency.}
This metric evaluates whether predicted claims are grounded in \emph{accessible} reference content.
For each generated claim $c \in \mathcal{C}^{\text{gen}}$, we first resolve its associated references
$\mathcal{R}^{\text{gen}}(c)$ into URLs, denoted as $\mathcal{R}^{\text{url}}(c)$.
We then retrieve their contents and define the subset of successfully accessed references as
$\mathcal{R}^{\text{acc}}(c) \subseteq \mathcal{R}^{\text{url}}(c)$.

We consider only claims with at least one resolvable URL:
\begin{equation}
\mathcal{C}^{\text{url}} = \{\, c \in \mathcal{C}^{\text{gen}} \mid |\mathcal{R}^{\text{url}}(c)| > 0 \,\}
\end{equation}

For each such claim, we define a binary verification function:
\begin{equation}
\mathrm{verify}(c) =
\begin{cases}
1, & |\mathcal{R}^{\text{acc}}(c)| > 0 \ \text{and} \ \mathrm{rel}(c, \mathcal{R}^{\text{acc}}(c)) = 1 \\
0, & \text{otherwise}
\end{cases}
\end{equation}
where $\mathrm{rel}(\cdot)$ is an LLM-based judgment that determines whether the claim and the retrieved content
are about the same or closely related specific topic.

The overall factual consistency score is defined as:
\begin{equation}
S_3(\hat{\mathcal{G}}) = \frac{1}{|\mathcal{C}^{\text{url}}|} \sum_{c \in \mathcal{C}^{\text{url}}} \mathrm{verify}(c)
\end{equation}

This metric captures the proportion of claims that are supported by accessible and topically consistent
external evidence, while penalizing missing, inaccessible, or irrelevant citations.

\textbf{Aggregate Score.}
The three metrics are combined into an overall evidence verification score:
\begin{equation}
S_{\mathrm{E}}(\hat{\mathcal{G}}) = \sum_{j=1}^{3} \alpha_j \cdot S_j(\hat{\mathcal{G}}), \quad \text{where} \quad \sum_{j=1}^{3} \alpha_j = 1
\end{equation}
Analogously to the holistic rubric, the evidence-verification weights $\alpha_j$ are task-adaptive, derived from the task specification and averaged over $T$ trials with normalization to $\sum_j \alpha_j = 1$. The final weights are fixed per task and shared across all systems (details in Appendix~\ref{app:benchmark_scale}).

\section{Performance Evaluation}
\label{sec:experiments}

\begin{table*}[!htb]
    \centering
    \small
    \caption{Overall performance of different models on MedProbeBench. The \textit{Overall Score} is a weighted composite metric. \textit{Holistic Quality} includes four medical-specific dimension scores (Compl.: Comprehensiveness, Insight: Insight Depth, Stnd.: Standards Adherence, Util.: Utility) and their composite score (Overall). \textit{Fine-grained Evidence Verification} includes three claim-level metrics (Rate: Task Success Rate, Effect.: Search Effectiveness, Fact.: Factual Consistency) and their composite score (Overall). Models are categorized by architecture: \textcolor{green!50!black}{LLM with Search Tools} (green) and \textcolor{orange!50!black}{Deep Research Agents} (orange).}
    \vspace{5mm}
    \resizebox{\textwidth}{!}{
    \begin{tabular}{lcccccccccc}
    \toprule
    \textbf{Model} & \textbf{Overall} & \multicolumn{5}{c}{\textbf{Holistic Quality}} & \multicolumn{4}{c}{\textbf{Fine-grained Evidence Verification}} \\
    \cmidrule(lr){3-7} \cmidrule(lr){8-11}
    & \textbf{Score$\uparrow$} & \textbf{Compl.$\uparrow$} & \textbf{Insight$\uparrow$} & \textbf{Stnd.$\uparrow$} & \textbf{Util.$\uparrow$} & \textbf{Overall$\uparrow$} & \textbf{Rate$\uparrow$} & \textbf{Effect.$\uparrow$} & \textbf{Fact.$\uparrow$} & \textbf{Overall$\uparrow$} \\
    \midrule
    \multicolumn{11}{l}{\textit{\textcolor{green!50!black}{LLM with Search Tools}}} \\
    \rowcolor{green!8} Claude-Sonnet 4                & 0.427 & 0.664 & 0.609 & 0.669 & 0.614 & 0.641 & 0.342 & 0.169 & 0.488 & 0.336 \\
    \rowcolor{green!8} Claude-Sonnet 4 (Think)        & 0.396 & 0.662 & 0.592 & 0.647 & 0.600 & 0.627 & 0.318 & 0.099 & 0.438 & 0.297 \\
    \rowcolor{green!8} Gemini-3-Flash                 & 0.484 & 0.700 & 0.650 & 0.712 & 0.639 & 0.678 & 0.319 & 0.354 & 0.669 & 0.401 \\
    \rowcolor{green!8} Gemini-3-Flash (Think)         & 0.464 & 0.683 & 0.629 & 0.705 & 0.627 & 0.664 & 0.305 & 0.339 & 0.611 & 0.378 \\
    \rowcolor{green!8} GPT-4.1                        & 0.445 & 0.666 & 0.625 & 0.682 & 0.632 & 0.652 & 0.301 & 0.288 & 0.570 & 0.356 \\
    \rowcolor{green!8} GPT-5                          & 0.472 & 0.711 & 0.689 & 0.737 & 0.648 & 0.700 & 0.340 & 0.296 & 0.544 & 0.374 \\
    \rowcolor{green!8} GPT-5.2                        & 0.527 & 0.745 & 0.728 & 0.772 & 0.682 & 0.736 & 0.353 & 0.405 & 0.693 & 0.437 \\
    \rowcolor{green!8} grok-4                         & 0.486 & 0.673 & 0.632 & 0.684 & 0.614 & 0.654 & 0.296 & 0.383 & 0.760 & 0.414 \\
    \rowcolor{green!8} Baichuan-M2-Plus                         & 0.529  & 0.710 & 0.656 & 0.697 & 0.654 & 0.681  & 0.338  & 0.311  & 0.951  & 0.463 \\
    \rowcolor{green!8} Baichuan-M3-Plus                         & 0.498  & 0.685  & 0.636 & 0.683 & 0.632 & 0.661   & 0.283  & 0.269  & \textbf{0.971} & 0.427 \\
    \midrule
    \multicolumn{11}{l}{\textit{\textcolor{orange!50!black}{Deep Research Agents}}} \\
    \rowcolor{orange!15} Tongyi DR-30B                  & 0.348 & 0.563 & 0.542 & 0.533 & 0.573 & 0.551 & 0.189 & 0.236 & 0.475 & 0.261 \\
    \rowcolor{orange!15} Kimi-K2.5-Agent & 0.540  & 0.730  & 0.668  & 0.714 & 0.653 & 0.695 & 0.415  & 0.415  & 0.689  & 0.474  \\
    \rowcolor{orange!15} AgentScope                     & 0.553 & 0.732 & 0.692 & 0.736 & 0.688 & 0.714 & 0.405 & 0.395 & 0.781 & 0.484 \\
    \rowcolor{orange!15} O4-mini DR                     & 0.561 & 0.726 & 0.673 & 0.738 & 0.641 & 0.699 & 0.444 & 0.254 & 0.908 & 0.502 \\
    \rowcolor{orange!15} Perplexity Sonar DR            & 0.576 & 0.699 & 0.650 & 0.704 & 0.615 & 0.672 & 0.447 & 0.450 & 0.857 & 0.536 \\
    \rowcolor{orange!15} MiroThinker-v1.5               & 0.614 & 0.780  & 0.743  & 0.779 & 0.695 & 0.754 & 0.433  & 0.483  & 0.946  & 0.554  \\
    \rowcolor{orange!15} MiroThinker-v1.5-pro              & \textbf{0.631} & \textbf{0.803}  & \textbf{0.757}  & \textbf{0.800} & \textbf{0.704} & \textbf{0.772} & \textbf{0.452}  & \textbf{0.496}  & 0.960  & \textbf{0.570}  \\
    \bottomrule
    \end{tabular}
    }
    \label{tab:overall}
\end{table*}

\subsection{Experimental Setup}

\textbf{Evaluated Systems.}
We evaluate 17 systems across three categories: 
(1) \textit{LLMs with Search Tools}: Claude Sonnet 4 in standard and thinking variants, Gemini 3 Flash Preview in standard and thinking variants, GPT-4.1, GPT-5, GPT-5.2, Grok-4, Baichuan-M2-Plus, and Baichuan-M3-Plus.
(2) \textit{Deep Research Agents, Commercial}: o4-mini-deep-research, Perplexity Sonar Deep Research~\citep{perplexity2025deepresearch}, and Kimi-K2.5-Agent~\citep{kimi_agent_2024}.
(3) \textit{Deep Research Agents, Open-source}: AgentScope-GPT4.1~\citep{gao2024agentscope}, MiroThinker-v1.5, MiroThinker-v1.5-pro~\citep{team2025mirothinker}, and Tongyi-DeepResearch-30B-A3B~\citep{team2025tongyi}.

\textbf{Evaluation Protocol.} The benchmark comprises 50 tasks, each corresponding to an individual tumor entity drawn from five WHO Classification volumes (9--12 tasks per domain; see Appendix~\ref{app:statistics}). Each system generates one complete guideline per task, yielding 50 guideline documents per system. For each task, systems are provided with the disease name, section schema, and output format specifications, and are required to generate complete guidelines without human intervention or retries. All systems are evaluated using zero-shot structured constraint prompting. For answer cleaning, we employ two preprocessing approaches for structured and unstructured outputs. Automated evaluation computes both holistic rubric scores and claim-level evidence verification metrics. We employ GPT-4.1~\citep{openai2024gpt4o} as the automated evaluator with fixed prompts and zero temperature, with reference retrieval through the JINA Reader API. The prompt templates used for system instruction and evaluation are in Appendix~\ref{app:template_prompts}.

Due to the high computational cost of deep research evaluation, each system is evaluated once per task. We discuss evaluation reliability in Appendix~\ref{app:eval_reliability}.

\subsection{Findings on Overall Performance}

\textbf{Conclusion 1: Deep Research Agents demonstrate architectural advantages over LLM+Search paradigms.}
Table~\ref{tab:overall} reveals a clear performance hierarchy across system architectures. Deep research agents occupy the top three positions: MiroThinker-v1.5, MiroThinker-v1.5-pro, Perplexity Sonar DR, demonstrating that multi-step research workflows with iterative retrieval and evidence aggregation provide substantial advantages for deep evidence integration in clinical guideline generation. 

Among LLM+Search baselines, Baichuan-M2-Plus achieves the strongest performance, approaching deep research performance. However, significant variance exists within deep research agent systems--Tongyi DR-30B underperforms most LLM+Search systems, suggesting that architectural design quality matters more than paradigm choice alone. Overall holistic scores are influenced jointly by both the underlying language model capability and the effectiveness of structured research workflows. 

\textbf{Insufficiency 1: Fine-grained evidence verification remains the primary bottleneck across all model architectures.}
Despite achieving moderate scores under the \emph{Holistic Rubrics}, all evaluated systems exhibit substantially weaker performance in \emph{Fine-grained Evidence Verification}, revealing a persistent gap between surface-level guideline quality and reliable evidence grounding.
Specifically, \emph{Task Success Rate} reveals incomplete coverage of expert-identified critical claims, while \emph{Search Effectiveness} exhibits consistently low performance across systems, suggesting difficulty in retrieving guideline-relevant sources. These patterns imply that current systems frequently rely on loosely related or indirect references rather than the most relevant evidence among retrieved documents.

Performance on \emph{Factual Consistency} varies across systems. Deep research systems exhibit relatively stronger claim--citation consistency when relevant evidence is available, whereas extended thinking modes do not consistently improve grounding performance and, in some cases, introduce additional inconsistencies. This suggests that increased reasoning depth alone is insufficient without reliable evidence retrieval and integration mechanisms.

\subsection{Findings on Knowledge Types}

\begin{figure}[!htb]
    \centering
    \includegraphics[width=\columnwidth]{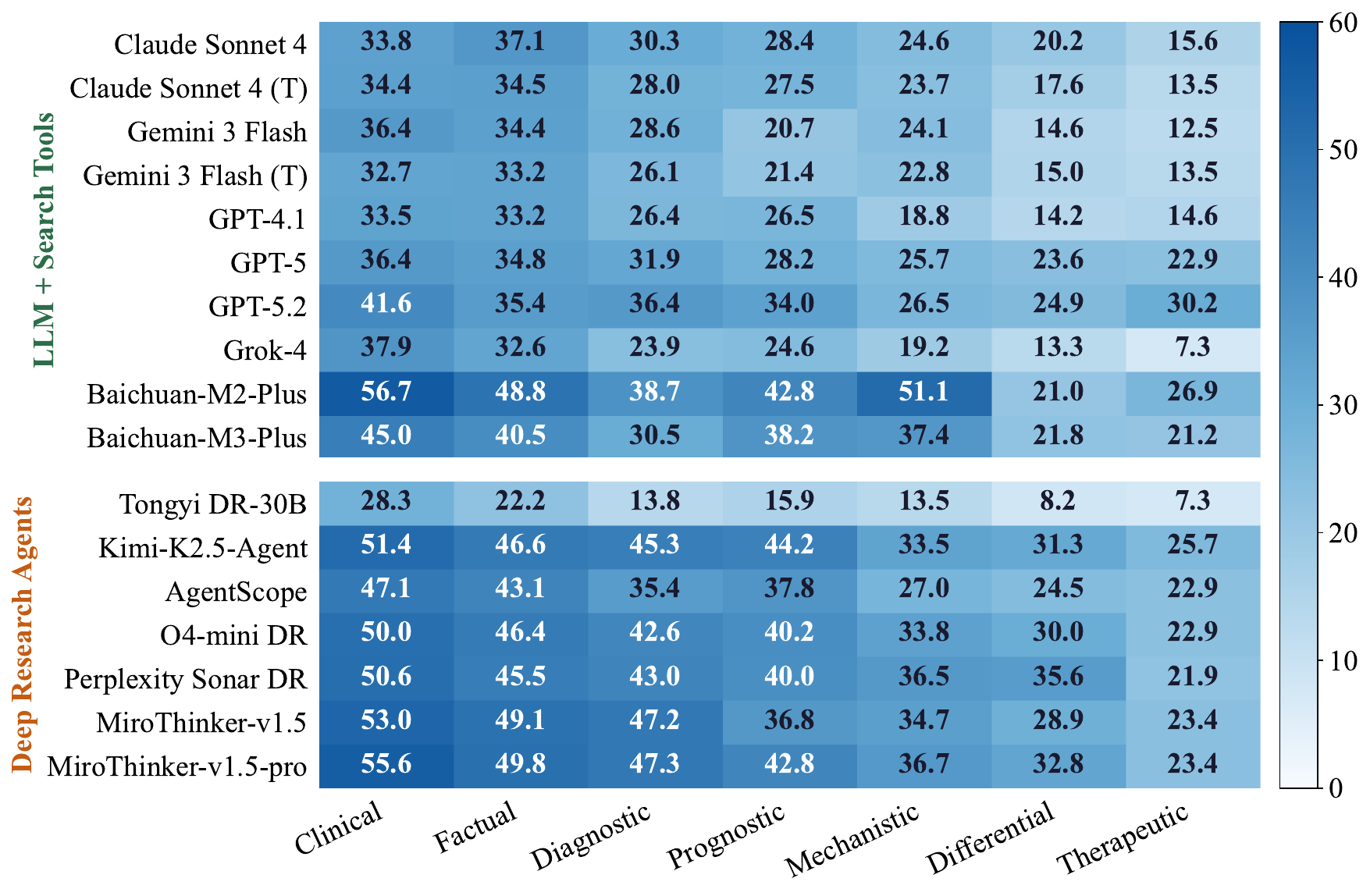}
    \caption{Task Success Rate (\%) by knowledge type across all evaluated systems. Color intensity reflects performance level; darker blue indicates higher coverage.}
    \label{fig:knowledge_heatmap}
    \end{figure}
    
    \begin{figure*}[!htb]
    \centering
    \includegraphics[width=\textwidth]{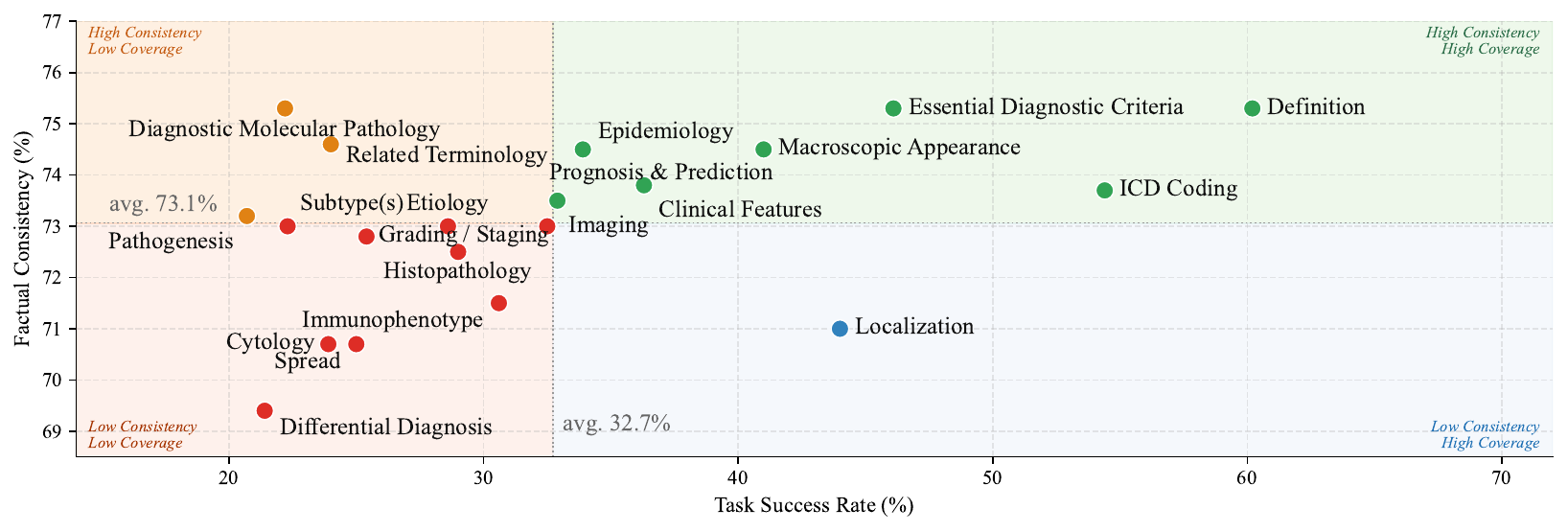}
    \caption{Coverage vs. Consistency by guideline section type. Each point represents one section type; color encodes coverage level. The pronounced horizontal spread against a narrow vertical range reveals that citation consistency remains uniformly high while claim coverage varies dramatically across sections, exposing a structural gap between surface fluency and substantive evidence grounding.}
    \label{fig:section_scatter}
    \end{figure*}

\textbf{Conclusion 2: Systems perform best on factual and clinical claims but struggle with specialized reasoning.}
Figure~\ref{fig:knowledge_heatmap} reveals significant performance variation across knowledge types. Systems achieve the highest accuracy on clinical and factual claims, indicating strong capabilities in descriptive knowledge, numerical data, and clinical practice standards. Diagnostic claims show moderate performance, while prognostic claims remain more challenging, confirming systems' usability in standard clinical documentation scenarios but limited robustness for outcome prediction. Top performer systems demonstrate substantial improvements: Baichuan-M2-Plus achieves 56.7\% accuracy on clinical claims, while mirothinker-v1.5-pro reaches 47.3\% on diagnostic claims, demonstrating that advanced architectures and specialized medical training substantially improve clinical knowledge integration.

\textbf{Insufficiency 2: Critical performance gaps in mechanistic reasoning and treatment guidance severely limit clinical utility.}
Performance drops sharply for knowledge types requiring deeper medical reasoning.
On average across all systems, mechanistic claims achieve only 28.8\% accuracy, differential diagnosis 22.2\%, and therapeutic claims 19.2\%, representing substantially lower performance than factual claims. However, notable progress has been made: Baichuan-M2-Plus achieves 51.1\% accuracy on mechanistic reasoning, demonstrating that specialized medical training can significantly improve performance on this critical dimension.

Mechanistic reasoning requires synthesizing molecular pathways, genotype-phenotype relationships, and pathophysiological processes, which are fundamental for precision medicine decisions (e.g., BRAF V600E mutations guiding targeted therapy). Despite improvements, therapeutic claim accuracy remains below 30\% even for the best systems, indicating persistent challenges in evidence-based treatment synthesis.
Overall, these results suggest that while recent advances show promise in mechanistic reasoning, current systems still face significant challenges in complex medical reasoning and evidence-based treatment synthesis, limiting their reliability in high-stakes clinical applications.

\subsection{Findings on Section Types}

\textbf{Conclusion 3: Citation consistency does not guarantee adequate evidence grounding.}
Figure~\ref{fig:section_scatter} reveals a clear decoupling between surface-level citation alignment and substantive evidence coverage. While content consistency remains uniformly high across guideline sections, claim coverage decreases sharply in sections that require specialized medical knowledge. 

Notably, diagnosis- and mechanism-oriented sections exhibit substantially lower factual coverage, despite maintaining comparable citation consistency. This pattern indicates that current systems can preserve structural fluency and citation formatting even when failing to retrieve or integrate the domain-specific evidence necessary to substantiate their claims, exposing a structural limitation in existing architectures.

\textbf{Insufficiency 3: Diagnostic reasoning failures undermine reliability for clinical decision support.}
The lowest performance is observed in sections that require comparative diagnostic reasoning rather than factual enumeration. Although systems recover established diagnostic criteria with moderate success, they struggle to distinguish between clinically similar conditions in differential diagnosis and to reason over underlying disease mechanisms. This failure reflects a limited capacity to perform exclusionary and comparative reasoning, which are core competencies in clinical diagnosis. As a result, these systems risk producing plausible but diagnostically misleading outputs, undermining their suitability for real-world clinical decision support where correct differentiation, rather than surface completeness, is critical.

\subsection{Reliability and Human-AI Alignment Validation}

\textbf{Human Evaluation Protocol.}  
To examine the alignment between automated evaluation and human expert judgment, we conduct a structured reader study using the same holistic rubric employed by the LLM-based evaluator. The four evaluation dimensions include medical coverage completeness, mechanistic insight depth, adherence to medical standards, and clinical utility. Each dimension is assessed based on multiple fine-grained criteria rather than a single coarse score, capturing clinically meaningful aspects of guideline quality such as workflow coverage, causal reasoning, standards compliance, and practical usability.

Human experts score each criterion on a 0–10 scale, and the resulting dimension scores are combined using the same task-level dynamic weights as the automated evaluation to obtain global scores. In particular, the dynamic weights are generated once from the task specification and section schema and then fixed across all systems and all human evaluations for that task, ensuring direct comparability between human and LLM-based assessments. We evaluate four representative systems across the performance spectrum in Table~\ref{tab:overall}, grouping MiroThinker-v1.5-pro and O4-mini DR as Good systems and Tongyi DR-30B and GPT-4.1 as Bad systems. Across five guideline categories, we sample two guidelines per system per category, resulting in 40 evaluation samples, each independently assessed by three senior medical experts.

\begin{figure}[!htb]
\centering
\includegraphics[width=1.0\columnwidth]{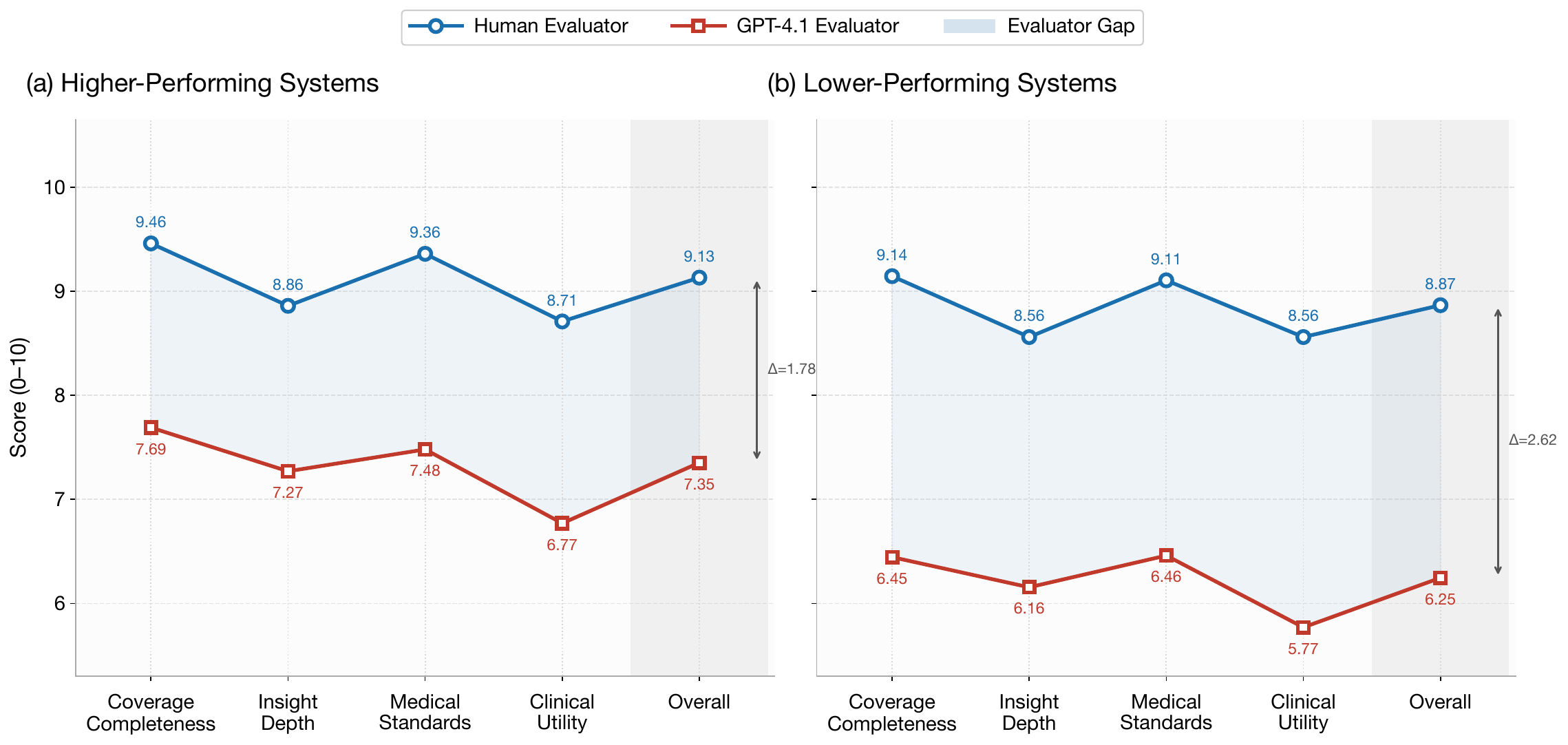}
\caption{Human and GPT-4.1 evaluation scores for higher-performing Good systems and lower-performing Bad systems across four dimensions and the overall score.}
\vspace{2mm}
\label{fig:human}
\end{figure} 

\textbf{Holistic Rubrics Alignment.}
Figure~\ref{fig:human} shows a clear and consistent alignment between human experts and the LLM judge across all evaluation dimensions. While GPT-4.1, used as the LLM judge, assigns systematically lower absolute scores than human evaluators for both high-quality (“Good”) and low-quality (“Bad”) guidelines, it preserves the same relative ordering and separation trends across dimensions. In particular, both evaluators consistently distinguish higher-quality guidelines from lower-quality ones, indicating strong discriminative capability of the automated judge. In contrast, human experts exhibit a generally higher scoring tendency, even for lower-quality outputs, suggesting that many generated guidelines meet a relatively strong baseline of clinical plausibility. Despite this difference in calibration, the parallel trends across dimensions indicate substantial agreement in relative quality judgments.
Quantitative analysis confirms statistically significant rank agreement between GPT-4.1 and human experts (Spearman $\rho = 0.754$, $p = 0.012$), with perfect directional agreement across all five dimensions. Detailed alignment statistics are reported in Table~\ref{tab:human_alignment} in Appendix~\ref{app:human_alignment}.

\textbf{Human Validation of Fine-grained Evidence Verification.}
We validate the automated factual consistency metric against human expert annotations on 200 stratified claim-reference pairs. The LLM judge achieves an F1 of $86.4\%$ with Cohen's $\kappa = 0.79$ (substantial agreement), confirming its reliability as an automated proxy for claim-level evidence grounding (details in Table~\ref{tab:evidence_human} in Appendix~\ref{app:human_alignment}).

\textbf{Dimension-Level Structural Insights.}
Across both human and automated evaluations, the four dimensions exhibit consistent and interpretable performance patterns. \textit{Medical Coverage Completeness} and \textit{Adherence to Medical Standards} receive the highest scores overall, indicating that current systems are generally proficient at covering core clinical content and adhering to widely accepted medical conventions. In contrast, \textit{Mechanistic Insight Depth} and \textit{Clinical Utility} show comparatively lower scores and larger gaps between high- and low-quality guidelines, highlighting these dimensions as primary sources of quality differences across systems. Notably, GPT-4.1 applies stricter penalties on mechanistic depth and clinical utility—particularly for lower-quality guidelines—suggesting increased sensitivity to shallow causal explanations and limited actionable reasoning. The close correspondence between human and automated score trends across all dimensions further supports the reliability of the automated judge in capturing fine-grained structural differences in guideline quality.

\section{Conclusion}
\label{sec:conclusion}

We introduced \textbf{MedProbeBench}, a benchmark for evaluating expert-level clinical guideline generation with claim-level evidence verification. Our results show that although state-of-the-art models generate fluent and well-structured guidelines, reliable evidence grounding remains a key bottleneck, especially for mechanistic reasoning and treatment-related content. High factual consistency often masks substantive failures in specialized knowledge integration, highlighting a gap between surface fluency and factual reliability. As medical AI systems move toward high-stakes clinical use, closing this synthesis–grounding gap is essential for trustworthy AI-assisted decision-making. MedProbeBench offers a standardized framework to expose these limitations and foster research on clinically reliable, evidence-based medical AI systems.

\clearpage
\newpage

\bibliographystyle{icml2025}
\bibliography{references}

\clearpage
\onecolumn
\appendix

\section{LLM-based Evaluation Protocol}
\label{app:prompt_template}

\subsection{Holistic Rubrics}

This prompt is used to evaluate the overall quality of a generated medical guideline by directly comparing it against a gold-standard reference guideline under a strict GT-anchored scoring protocol.
For each task, before scoring any system outputs, the same judge model is additionally prompted to produce task-specific weights over the four fixed holistic dimensions based only on the disease specification and required section schema, following a task-adaptive weighting strategy inspired by DeepResearch Bench~\citep{du2025deepresearch}. The weight generation step is repeated for $T$ trials and averaged, and the resulting task-level weights are then shared across all systems for that task.

\begin{lstlisting}
You are a senior medical guideline reviewer conducting a STRICT BENCHMARK evaluation.

Your task is to score a GENERATED medical guideline by DIRECTLY COMPARING it to a REFERENCE guideline (ground truth).

## Reference Guideline (Ground Truth, Score = 10):
{reference_content}

## Generated Guideline (To Be Scored by Deviation):
{generated_content}

## Evaluation Criteria (Each scored independently, GT = 10):
{criteria_list}

IMPORTANT ANCHOR RULE:
- The REFERENCE guideline represents perfect, expert-level quality and is implicitly scored as 10/10.
- The GENERATED guideline is scored ONLY by how much it DEVIATES from the reference.
- Scores MUST be assigned by starting from 10 and subtracting points based on differences, weaknesses, or omissions.

## Relative Scoring Philosophy (GT-ANCHORED):
- 10: Generated guideline is effectively identical to the reference (or equivalent in all clinically meaningful content)
- 8-9: Minor deviations from the reference (small omissions, reduced detail, or wording differences without clinical impact)
- 6-7: Noticeable deviations (missing rationale, reduced specificity, or weaker guidance in several areas)
- 4-5: Major deviations (important sections missing, oversimplified guidance, or reduced clinical safety)
- 2-3: Severe deviations (critical content missing, inaccuracies, or potentially unsafe recommendations)
- 0-1: Extreme deviation (fundamentally incorrect, misleading, or largely unrelated to the reference)

## Mandatory Benchmark Rules (NON-NEGOTIABLE):
1. GT = 10 by definition - do NOT independently judge quality
2. Score ONLY the distance between GENERATED and REFERENCE
3. Any missing, weaker, or less specific content MUST reduce the score
4. Vagueness counts as deviation when the reference is specific or evidence-based
5. Equal section titles do NOT imply equivalence - content depth and precision matter
6. When uncertain, subtract points - never inflate scores in benchmark settings

## Output Requirements (STRICT):
- Output ONLY a valid JSON object
- No explanations, no markdown, no additional text
- Use EXACTLY the following structure:

{"scores": [
  {"criterion": "<criterion name>", "score": <integer 0-10>, "reason": "<explicit difference from GT explaining point deductions>"}
]}
## Example (FORMAT ONLY):
{"scores": [
  {"criterion": "Clinical Specificity", "score": 7, "reason": "GT specifies exact thresholds and dosages; generated omits numerical criteria"},
  {"criterion": "Safety Considerations", "score": 5, "reason": "GT details contraindications and monitoring; generated mentions safety only in general terms"}
]}
\end{lstlisting}

\subsection{Fine-grained Evidence Verification}

This section details the full fine-grained evidence verification pipeline. We first describe the claim matching threshold used in semantic matching (\S\ref{app:claim_matching}), then the claim extraction prompt that decomposes guidelines into atomic claims (\S\ref{app:claim_extraction}). We then present three independent prompt templates for evaluating evidence grounding at the claim and reference level, corresponding to task success rate (\S\ref{app:task_rate}), search effectiveness (\S\ref{app:search_effectiveness}), and factual consistency (\S\ref{app:factual_consistency}).

\subsubsection{Claim Matching Threshold}
\label{app:claim_matching}

In the two-stage claim matching procedure, we use a Jaccard similarity threshold $\theta=0.85$ as a conservative pre-filter. Claims with a Jaccard score at or above $\theta$ are considered matched directly; all remaining pairs are passed to an LLM-based semantic evaluator. This threshold is intentionally set high so that only claims with exceptionally high lexical overlap bypass LLM evaluation, ensuring that all borderline cases receive semantic judgment. Since the actual matching decision for most claim pairs is made by the LLM, the specific value of $\theta$ has minimal impact on final evaluation quality.

\subsubsection{Claim Extraction}
\label{app:claim_extraction}

This prompt is applied to each normalized section of a guideline (both gold-standard and generated) to extract atomic, independently verifiable clinical claims. Each claim is returned as a structured record containing the claim text, associated citation markers, and a knowledge-type label.

\begin{lstlisting}
# Medical Claim Extraction

## Task
Extract medical claims from the following medical guideline section.
Each claim must be a minimal, independently verifiable factual statement.

## Section Text
{section_text}

## Output Format
Return a JSON object with the following structure:
{
  "claims": [
    {
      "id": "C001",
      "content": "Extracted factual statement",
      "reference": "[1]",
      "type_knowledge": "Factual",
      "section": "<section name>"
    }
  ]
}

## Field Definitions
- id: Sequential identifier (C001, C002, ...)
- content: Claim text (preserve original wording, values, and annotations)
- reference: In-text citation markers such as "[1]", "[2, 3]", or "" if absent
- type_knowledge: One of [Factual, Mechanistic, Clinical, Diagnostic,
                          Differential, Prognostic, Therapeutic]
- section: The section name from which the claim was extracted

## Knowledge Type Guide
- Factual: Descriptive facts, statistics, anatomical locations, ICD codes
- Mechanistic: Causal mechanisms, molecular pathways, genetic alterations
- Clinical: Clinical manifestations, signs, symptoms
- Diagnostic: Diagnostic tests, imaging or pathology findings, criteria
- Differential: Features distinguishing a disease from related conditions
- Prognostic: Survival statistics, outcome predictors, recurrence rates
- Therapeutic: Treatment methods, drug regimens, efficacy or toxicity data

## Extraction Rules
1. One fact per claim: split compound sentences into atomic units
2. Preserve details: retain all quantitative values, percentages, and ranges
3. Preserve aliases: e.g., "SMARCB1 (also known as hSNF5, INI1, or BAF47)"
4. Omit figure and table cross-references (e.g., "see Figure 1")
5. Do not paraphrase: keep the original medical terminology
\end{lstlisting}

\subsubsection{Task Success Rate}
\label{app:task_rate}

This prompt determines whether a gold-standard (GT) claim is semantically covered by any predicted claim.

\begin{lstlisting}
Determine if the GT claim is semantically covered by any predicted claim.

GT Claim: {gt_claim}

Predicted Claims:
{pred_claims_json}

Match if: same medical fact, equivalent meaning, or prediction covers GT's core info.

Return JSON only:
{"is_hit": true/false, "matched_pred_claim_id": "id or empty"}
\end{lstlisting}

\subsubsection{Search Effectiveness}
\label{app:search_effectiveness}

This prompt evaluates whether a predicted reference corresponds to the same underlying object as the gold-standard reference.

\begin{lstlisting}
Determine whether the following two URLs refer to the same underlying topic, work, or object.

GT URL: {gt_url}
GT Content (excerpt):
{gt_content}

Pred URL: {pred_url}
Pred Content (excerpt):
{pred_content}

Consider it a match if both contents discuss the same underlying thing (e.g., the same study, dataset, method, system, or phenomenon), even if their wording, focus, conclusions, or level of detail differ.

If one or both contents are mainly meta-information (e.g., titles, abstracts, or repository pages), use available clues (such as names or referenced objects) and apply a slightly more permissive judgment.

Ignore minor differences or disagreements.

Return JSON only:
{"is_match": true/false, "reason": "brief explanation of the shared underlying object"}
\end{lstlisting}

\subsubsection{Factual Consistency}
\label{app:factual_consistency}

This prompt assesses whether a generated claim is actually supported by its cited reference content---that is, whether the specific factual assertion made in the claim can be substantiated by the evidence in the reference, rather than merely sharing a topical area.

\begin{lstlisting}
Determine whether the cited reference content provides evidence that supports the specific claim.

Claim:
{claim}

Reference Content:
{reference_content}

Your task is to judge evidential support, not topical overlap.

Return TRUE only if the reference content contains information that substantiates the specific factual assertion in the claim.
This includes: numerical values, diagnostic criteria, molecular findings, prevalence data, prognosis estimates, or mechanistic relationships that are stated or clearly implied in the reference.

Return TRUE if the reference is only an abstract or brief metadata, but the abstract explicitly reports findings consistent with the claim (e.g., the abstract reports the same gene mutation, the same survival statistic, or the same diagnostic criterion stated in the claim).

Return FALSE if:
- The claim makes a specific factual assertion (e.g., a percentage, a molecular marker, a treatment outcome) that is not mentioned or supported in the reference content.
- The reference only covers the same broad disease area or topic without providing evidence for the specific claim.
- The reference contradicts or is inconsistent with the claim.
- The reference is purely meta-information (title/landing page only) with no substantive findings that match the claim.

Return JSON only:
{"is_consistent": true/false}
\end{lstlisting}

\section{Guideline Generation Prompt Template}
\label{app:template_prompts}

Below we present the prompt template used for medical guideline generation tasks. The template enforces structured output, evidence usage constraints, and explicit citation requirements.

\begin{lstlisting}
# System Role
{role_and_constraints}

# Task
Write a comprehensive medical guideline chapter for: **{tumor_name}**

# Evidence & Literature Guidance
- You are encouraged to leverage scientific literature (e.g., PubMed-indexed studies) to enhance the accuracy, depth, and credibility of the content.
- When relevant, distinguish evidence derived from:
  - Randomized controlled trials (RCTs)
  - Cohort or case-control studies
  - Exploratory or molecular/mechanistic studies
- Clearly reflect the strength and limitations of available evidence.

# Literature Usage Constraints
- Only cite articles with verifiable PubMed IDs or DOIs. Do not fabricate references.
- ONLY search for original research papers published in peer-reviewed journals.

# Citation Rules
- Every factual, clinical, or scientific claim must be supported by at least one appropriate citation, unless explicitly stated as expert opinion or lacking available evidence.
- Use inline numeric citations in square brackets, e.g., [1], [2], or [1, 2, 3].

# Required Sections
{sections_text}

# Output Format
Output in Markdown format:

# {tumor_name}

## Section Name
Structured, guideline-style content with clear clinical relevance.
Evidence-based discussion with inline citations.
Key statistics, outcomes, and limitations where applicable...

...(Continue for all sections - remember citations in EVERY section)

## References
1. Author et al. Title. *Journal*. Year;Volume:Pages. URL: https://pubmed.ncbi.nlm.nih.gov/XXXXXXXX/
2. Author et al. Title. *Journal*. Year;Volume:Pages. URL: https://doi.org/10.XXXX/XXXXX

...(List all cited references)
\end{lstlisting}



\section{Evaluation Reliability}
\label{app:eval_reliability}

Due to the high computational cost of deep research evaluation, each system is evaluated once per task. Result reliability is supported by three factors: (1) each system's score aggregates over approximately 5{,}130 claim-level judgments across 50 tasks, providing fine-grained statistical coverage beyond task-level counts; (2) performance rankings remain consistent across all five medical domains (see Appendix~\ref{app:statistics}), indicating stable cross-domain conclusions; and (3) single-round evaluation is standard practice in comparable long-form research benchmarks~\citep{du2025deepresearch}.

\section{Human--AI Alignment Analysis}
\label{app:human_alignment}

This section reports the detailed quantitative results underlying the human--AI alignment validation presented in Section~\ref{sec:experiments}.
We validate both tiers of MedProbeBench's evaluation framework: holistic rubric scoring (\S\ref{app:holistic_alignment}) and fine-grained evidence verification (\S\ref{app:evidence_alignment}).
In each case, we summarize the study design, report agreement statistics, and discuss implications for the reliability of automated evaluation.

\subsection{Holistic Rubric Alignment}
\label{app:holistic_alignment}

Following the evaluation protocol described in Section~\ref{sec:experiments}, three senior medical experts independently evaluate 40 guideline samples (4 systems $\times$ 5 categories $\times$ 2 guidelines) using the same four-dimensional rubric and task-level dynamic weights as the automated GPT-4.1 judge.
To quantify agreement, we compute Spearman rank correlation between GPT-4.1 scores and aggregated human expert scores across all dimension--group combinations. Table~\ref{tab:human_alignment} summarizes the results.

\begin{table}[h]
\centering
\small
\caption{Human--AI alignment statistics for holistic rubric evaluation. Scores are on a 0--10 scale.}
\label{tab:human_alignment}
\begin{tabular}{lr}
\toprule
\textbf{Metric} & \textbf{Value} \\
\midrule
Spearman $\rho$ & 0.754 \\
$p$-value & 0.012 \\
Directional agreement & 5/5 dimensions \\
Mean absolute difference & 2.21 \\
\bottomrule
\end{tabular}
\end{table}

Both evaluators consistently rank higher-quality systems above lower-quality ones across all five evaluation dimensions, yielding perfect directional agreement. The mean absolute scoring difference of 2.21 points reflects a systematic calibration offset rather than disagreement in discriminative judgment: human experts tend to assign higher absolute scores, but the relative ranking of systems is preserved. These results collectively validate GPT-4.1 as a reliable proxy for expert judgment in holistic quality assessment.

\subsection{Fine-grained Evidence Verification Alignment}
\label{app:evidence_alignment}

As described in Section~\ref{sec:experiments}, we conduct a targeted human validation study to assess the reliability of LLM-based factual consistency verification. We randomly sample 200 claim-reference pairs from the full evaluation set, stratified across section types and model outputs to ensure broad coverage. Three senior medical experts independently annotate each pair as \textit{supported} or \textit{not supported}, based on whether the cited reference genuinely substantiates the associated clinical claim; majority vote serves as the gold standard. Results are summarized in Table~\ref{tab:evidence_human}.

\begin{table}[h]
\centering
\small
\caption{Human validation of automated factual consistency verification. Metrics are computed against majority-vote human expert labels on 200 stratified claim-reference pairs.}
\label{tab:evidence_human}
\begin{tabular}{lr}
\toprule
\textbf{Metric} & \textbf{Value} \\
\midrule
Precision & 87.3\% \\
Recall & 85.6\% \\
F1 & 86.4\% \\
Cohen's $\kappa$ & 0.79 \\
\bottomrule
\end{tabular}
\end{table}

These results confirm that GPT-4.1-based factual consistency verification reliably aligns with expert judgment, validating its use as an automated proxy for claim-level evidence grounding.

\section{Dataset Statistics}
\label{app:statistics}

This section provides detailed statistics on the MedProbeBench dataset and supplementary evaluation breakdowns referenced from the main text.
We first describe the ground-truth data composition: claim type distribution (\S\ref{app:claim_type_dist}), dataset composition by medical domain (\S\ref{app:domain_comp}), and benchmark scale statistics (\S\ref{app:benchmark_scale}).
We then report fine-grained evaluation breakdowns that supplement the findings in Section~\ref{sec:experiments}: task success rate by knowledge type (\S\ref{app:tab_knowledge}) and system performance by section type (\S\ref{app:tab_section}).

\subsection{Claim Type Distribution}
\label{app:claim_type_dist}
We decompose all gold-standard guidelines into atomic claims following the methodology described in Section~\ref{sec:rubric}.
Each claim is categorized into one of seven knowledge types according to its semantic content and clinical role.
Table~\ref{tab:claim_types} summarizes the distribution of knowledge types across the entire benchmark.

\begin{table}[h]
\centering
\small
\begin{tabular}{lcc}
\toprule
\textbf{Knowledge Type} & \textbf{Count} & \textbf{Percentage} \\
\midrule
Clinical (Clin.) & 379 & 7.38\% \\
Factual (Fact.) & 2422 & 47.18\% \\
Diagnostic (Diag.) & 1060 & 20.65\% \\
Prognostic (Prog.) & 443 & 8.63\% \\
Mechanistic (Mech.) & 482 & 9.39\% \\
Differential (Diff.) & 247 & 4.81\% \\
Therapeutic (Ther.) & 100 & 1.95\% \\
\midrule
\textbf{Total} & \textbf{5133} & \textbf{100\%} \\
\bottomrule
\end{tabular}
\caption{Distribution of knowledge types in MedProbeBench.}
\label{tab:claim_types}
\end{table}

\subsection{Dataset Composition by Medical Domain}
\label{app:domain_comp}

The benchmark encompasses five WHO classification volumes, covering diverse organ systems and disease categories. Table~\ref{tab:domain_dist} summarizes the distribution of guidelines and claims across different medical domains.

\begin{table}[h]
\centering
\small
\begin{tabular}{lccc}
\toprule
\textbf{Medical Domain} & \textbf{Guidelines} & \textbf{Claims} & \textbf{Avg. Claims/Guideline} \\
\midrule
Central Nervous System Tumours & 9 & 970 & 107.78 \\
Haematolymphoid Tumours & 12 & 1033 & 86.08 \\
Digestive System Tumours & 9 & 1213 & 134.78 \\
Thoracic Tumours & 10 & 1181 & 118.10 \\
Soft Tissue and Bone Tumours & 10 & 736 & 73.60 \\
\midrule
\textbf{Total} & \textbf{50} & \textbf{5133} & \textbf{102.66} \\
\bottomrule
\end{tabular}
\caption{Dataset composition by medical domain.}
\label{tab:domain_dist}
\end{table}

\subsection{Benchmark Scale Statistics}
\label{app:benchmark_scale}

Table~\ref{tab:rubric_stats} reports the full scale of MedProbeBench's evaluation infrastructure: the number of task-adaptive holistic rubric criteria generated across all 50 tasks, broken down by dimension, alongside the citation coverage of the benchmark's ground-truth claims.

\paragraph{Task-Adaptive Weight Derivation.}
For holistic rubric evaluation, we derive task-specific dimension weights $w_i$ from the task specification and required section schema, while keeping the four top-level medical dimensions fixed. For each task $t$, the judge LLM independently produces dimension weights $w_i^{(m,t)}$ over $T$ trials, and we use the averaged weights:
\begin{equation}
w_i^{(t)} = \frac{1}{T} \sum_{m=1}^{T} w_i^{(m,t)}, \quad i \in \{1,2,3,4\},
\end{equation}
followed by normalization so that $\sum_i w_i^{(t)} = 1$. The resulting weight vector is generated once per task and then shared by all evaluated systems on that task, ensuring consistent cross-system comparison.

Analogously, for evidence verification, we assign task-specific weights $\alpha_j$ to the three metrics based on the same task description and section schema. For each task $t$, these weights are averaged across $T$ independent trials:
\begin{equation}
\alpha_j^{(t)} = \frac{1}{T} \sum_{m=1}^{T} \alpha_j^{(m,t)}, \quad j \in \{1,2,3\},
\end{equation}
and normalized to satisfy $\sum_j \alpha_j^{(t)} = 1$. As with $w_i^{(t)}$, the final $\alpha_j^{(t)}$ are fixed at the task level and reused for all systems evaluated on that task.

\begin{table}[h]
\centering
\small
\caption{Benchmark scale statistics for MedProbeBench. Holistic rubric criteria are generated per-task and summed across all 50 tasks. Claims with references are GT claims carrying at least one supporting citation.}
\label{tab:rubric_stats}
\begin{tabular}{lrr}
\toprule
\textbf{Component} & \textbf{Count} & \textbf{Details} \\
\midrule
\multicolumn{3}{l}{\textit{Holistic Rubric Criteria (across 50 tasks)}} \\
\quad Comprehensiveness   & 352  & \\
\quad Insight Depth       & 296  & \\
\quad Accuracy \& Standards & 294 & \\
\quad Clinical Utility    & 294  & \\
\quad \textbf{Total Rubric Criteria} & \textbf{1,236} & $\approx$24.7 per task \\
\midrule
\multicolumn{3}{l}{\textit{Ground-Truth Claim Statistics}} \\
\quad Total atomic claims & 5,133 & across 50 tasks \\
\quad Claims with $\geq$1 citation & 2,203 & 42.92\% of total \\
\quad Claims without citation & 2,930 & 57.08\% of total \\
\bottomrule
\end{tabular}
\end{table}

The 1,236 total rubric criteria underscore the fine-grained, disease-specific nature of MedProbeBench's holistic evaluation---each task receives its own rubric tailored to the disease's clinical priorities, rather than a single fixed template applied uniformly.
The 42.92\% citation rate in GT claims reflects the evidence grounding standard of WHO guidelines: highly descriptive or definitional claims (terminology, ICD codes, macroscopic features) typically carry no citation, while mechanistic, prognostic, and diagnostic claims are expected to be evidence-backed.

\subsection{Task Success Rate by Knowledge Type}
\label{app:tab_knowledge}
Table~\ref{tab:type_rate_app} reports the full task success rate breakdown by knowledge type for all evaluated systems, supplementing the analysis in Section~\ref{sec:experiments} (Findings on Knowledge Types).

\begin{table}[h]
\centering
\scriptsize
\resizebox{0.7\columnwidth}{!}{%
\begin{tabular}{lcccccccc}
\toprule
\textbf{Model} & \textbf{Clin.} & \textbf{Fact.} & \textbf{Diag.} & \textbf{Prog.} & \textbf{Mech.} & \textbf{Diff.} & \textbf{Ther.} \\
\midrule
\multicolumn{8}{l}{\textit{\textcolor{green!50!black}{LLM with Search Tools}}} \\
\rowcolor{green!8} Claude Sonnet 4           & 33.8 & 37.1 & 30.3 & 28.4 & 24.6 & 20.2 & 15.6 \\
\rowcolor{green!8} Claude Sonnet 4 (T)       & 34.4 & 34.5 & 28.0 & 27.5 & 23.7 & 17.6 & 13.5 \\
\rowcolor{green!8} Gemini 3 Flash            & 36.4 & 34.4 & 28.6 & 20.7 & 24.1 & 14.6 & 12.5 \\
\rowcolor{green!8} Gemini 3 Flash (T)        & 32.7 & 33.2 & 26.1 & 21.4 & 22.8 & 15.0 & 13.5 \\
\rowcolor{green!8} GPT-4.1                   & 33.5 & 33.2 & 26.4 & 26.5 & 18.8 & 14.2 & 14.6 \\
\rowcolor{green!8} GPT-5                     & 36.4 & 34.8 & 31.9 & 28.2 & 25.7 & 23.6 & 22.9 \\
\rowcolor{green!8} GPT-5.2                   & 41.6 & 35.4 & 36.4 & 34.0 & 26.5 & 24.9 & \textbf{30.2} \\
\rowcolor{green!8} grok-4                    & 37.9 & 32.6 & 23.9 & 24.6 & 19.2 & 13.3 & 7.3 \\
\rowcolor{green!8} Baichuan-M2-Plus          & \textbf{56.7} & 48.8 & 38.7 & 42.8 & \textbf{51.1} & 21.0 & 26.9 \\
\rowcolor{green!8} Baichuan-M3-Plus          & 45.0 & 40.5 & 30.5 & 38.2 & 37.4 & 21.8 & 21.2 \\
\midrule
\multicolumn{8}{l}{\textit{\textcolor{orange!50!black}{Deep Research Agents}}} \\
\rowcolor{orange!15} AgentScope                & 47.1 & 43.1 & 35.4 & 37.8 & 27.0 & 24.5 & 22.9 \\
\rowcolor{orange!15} Tongyi DR-30B             & 28.3 & 22.2 & 13.8 & 15.9 & 13.5 & 8.2 & 7.3 \\
\rowcolor{orange!15} O4-mini DR                & 50.0 & 46.4 & 42.6 & 40.2 & 33.8 & 30.0 & 22.9 \\
\rowcolor{orange!15} Perplexity Sonar DR       & 50.6 & 45.5 & 43.0 & 40.0 & 36.5 & \textbf{35.6} & 21.9 \\
\rowcolor{orange!15} kimi-agent                & 51.4 & 46.6 & 45.3 & \textbf{44.2} & 33.5 & 31.3 & 25.7 \\
\rowcolor{orange!15} mirothinker-v1.5          & 53.0 & 49.1 & 47.2 & 36.8 & 34.7 & 28.9 & 23.4 \\
\rowcolor{orange!15} mirothinker-v1.5-pro      & 55.6 & \textbf{49.8} & \textbf{47.3} & 42.8 & 36.7 & 32.8 & 23.4 \\
\midrule
Type Average & 42.6 & 39.2 & 33.8 & 32.4 & 28.8 & 22.2 & 19.2 \\
\bottomrule
\end{tabular}}
\caption{Task Success Rate by knowledge type (\%).}
\label{tab:type_rate_app}
\end{table}

\subsection{System Performance by Section Type}
\label{app:tab_section}
Table~\ref{tab:section_rate_app} reports task success rate and factual consistency across all 20 guideline section types, averaged over all systems, supplementing the analysis in Section~\ref{sec:experiments} (Findings on Section Types).

\begin{table}[h]
\centering
\small
\resizebox{\columnwidth}{!}{%
\begin{tabular}{lcc|lcc}
\toprule
\textbf{Section Type} & \textbf{Task Success Rate$\uparrow$} & \textbf{Consistency$\uparrow$} & \textbf{Section Type} & \textbf{Task Success Rate$\uparrow$} & \textbf{Consistency$\uparrow$} \\
\midrule
Definition & \textbf{60.2\%} & 75.3\% & Epidemiology & 33.9\% & 74.5\% \\
ICD Coding & 54.4\% & 73.7\% & Prognosis and prediction & 32.9\% & 73.5\% \\
Essential diagnostic criteria & 46.1\% & 75.3\% & Imaging & 32.5\% & 73.0\% \\
Localization & 44.0\% & 71.0\% & Immunophenotype & 30.6\% & 71.5\% \\
Macroscopic appearance & 41.0\% & 74.5\% & Histopathology & 29.0\% & 72.5\% \\
Clinical features & 36.3\% & 73.8\% & Etiology & 28.6\% & 73.0\% \\
\midrule
Grading / Staging & 25.4\% & 72.8\% & Cytology & 23.9\% & 70.7\% \\
Spread & 25.0\% & 70.7\% & Subtype(s) & 22.3\% & 73.0\% \\
Diagnostic molecular pathology & 22.2\% & 75.3\% & Differential diagnosis & 21.4\% & 69.4\% \\
Pathogenesis & 20.7\% & 73.2\% & Related terminology & 24.0\% & 74.6\% \\
\midrule
Section Average & 32.7\% & 73.1\% & & & \\
\bottomrule
\end{tabular}}
\caption{System performance by guideline section type. Task success rate measures claim coverage accuracy; Consistency measures citation-claim alignment.}
\label{tab:section_rate_app}
\end{table}

\section{Multi-Judge Validation}
\label{app:multi_judge}

To examine potential self-preference bias and confirm the reliability of GPT-4.1 as the automated judge, we conduct a multi-judge validation study using three independent LLM judges: GPT-4.1, Claude-4-Sonnet, and Gemini-2.5-Pro.
We randomly sample 200 guideline evaluations drawn from diverse models and disease topics across the benchmark, and each judge independently scores all samples on both Holistic Quality and Fine-grained Evidence Verification.

Tables~\ref{tab:multijudge_holistic} and~\ref{tab:multijudge_evidence} report per-model scores from each judge, together with Intraclass Correlation Coefficient (ICC) and Mean Absolute Deviation (MAD) between GPT-4.1 and the other two judges.
Across both evaluation dimensions, score differences between judges remain within 0.01--0.03, and all ICC values exceed 0.96, indicating excellent inter-judge agreement.
Notably, GPT-4.1 does not assign systematically higher scores to GPT-family models (GPT-5, GPT-5.2) relative to scores from Claude or Gemini judges, and Claude-4-Sonnet does not receive inflated scores from the Claude judge, confirming the absence of self-preference bias.
These results validate GPT-4.1 as a reliable and unbiased automated evaluator for MedProbeBench.

\begin{table}[h]
\centering
\small
\caption{Multi-judge validation on Holistic Quality. Scores are overall holistic quality composites (0--1 scale). MAD is computed between GPT-4.1 and each alternative judge across the 200 sampled evaluations.}
\label{tab:multijudge_holistic}
\begin{tabular}{lccc|cc}
\toprule
\textbf{Model} & \textbf{GPT-4.1} & \textbf{Claude} & \textbf{Gemini} & \textbf{MAD\textsubscript{Claude}} & \textbf{MAD\textsubscript{Gemini}} \\
\midrule
MiroThinker-v1.5-pro  & 0.772 & 0.768 & 0.775 & 0.004 & 0.003 \\
MiroThinker-v1.5      & 0.754 & 0.751 & 0.758 & 0.003 & 0.004 \\
AgentScope            & 0.714 & 0.720 & 0.718 & 0.006 & 0.004 \\
GPT-5.2               & 0.736 & 0.741 & 0.739 & 0.005 & 0.003 \\
GPT-5                 & 0.700 & 0.697 & 0.704 & 0.003 & 0.004 \\
Perplexity Sonar DR   & 0.672 & 0.669 & 0.675 & 0.003 & 0.003 \\
Claude-Sonnet 4       & 0.641 & 0.638 & 0.645 & 0.003 & 0.004 \\
Tongyi DR-30B         & 0.551 & 0.547 & 0.553 & 0.004 & 0.002 \\
\midrule
\multicolumn{4}{l}{ICC (95\% CI)} & \multicolumn{2}{c}{0.971~~[0.965,~0.977]} \\
\multicolumn{4}{l}{Overall MAD}   & 0.004 & 0.003 \\
\bottomrule
\end{tabular}
\end{table}

\begin{table}[h]
\centering
\small
\caption{Multi-judge validation on Fine-grained Evidence Verification. Scores are overall evidence verification composites (0--1 scale).}
\label{tab:multijudge_evidence}
\begin{tabular}{lccc|cc}
\toprule
\textbf{Model} & \textbf{GPT-4.1} & \textbf{Claude} & \textbf{Gemini} & \textbf{MAD\textsubscript{Claude}} & \textbf{MAD\textsubscript{Gemini}} \\
\midrule
MiroThinker-v1.5-pro  & 0.570 & 0.563 & 0.574 & 0.007 & 0.004 \\
MiroThinker-v1.5      & 0.554 & 0.548 & 0.560 & 0.006 & 0.006 \\
AgentScope            & 0.484 & 0.479 & 0.491 & 0.005 & 0.007 \\
GPT-5.2               & 0.437 & 0.431 & 0.443 & 0.006 & 0.006 \\
GPT-5                 & 0.374 & 0.369 & 0.378 & 0.005 & 0.004 \\
Perplexity Sonar DR   & 0.536 & 0.529 & 0.541 & 0.007 & 0.005 \\
Claude-4-Sonnet 4       & 0.336 & 0.341 & 0.332 & 0.005 & 0.004 \\
Tongyi DR-30B         & 0.261 & 0.257 & 0.265 & 0.004 & 0.004 \\
\midrule
\multicolumn{4}{l}{ICC (95\% CI)} & \multicolumn{2}{c}{0.963~~[0.955,~0.971]} \\
\multicolumn{4}{l}{Overall MAD}   & 0.006 & 0.005 \\
\bottomrule
\end{tabular}
\end{table}

\section{End-to-End Evaluation Walkthrough}
\label{app:walkthrough}

To address reviewer concerns about the lack of concrete evaluation examples, we present a complete end-to-end walkthrough of the MedProbeBench evaluation pipeline using a representative case: \textit{Atypical Spindle Cell/Pleomorphic Lipomatous Tumour} (ASPLT).
This entity was selected for its compact yet structurally complete guideline (20 sections, 56 ground-truth claims, 15 references), making it suitable for detailed illustration without overwhelming length.

\subsection{Case Overview}

\begin{table}[h]
\centering
\small
\caption{Comparison of ground-truth (GT) and GPT-4.1 prediction statistics for the ASPLT case.}
\label{tab:asplt_overview}
\begin{tabular}{lcc}
\toprule
\textbf{Property} & \textbf{GT (Gold Standard)} & \textbf{GPT-4.1 Prediction} \\
\midrule
Number of sections       & 20    & 20 \\
Total claims extracted   & 56    & 76 \\
Claims with references   & 24 (42.9\%) & 33 (43.4\%) \\
Unique reference entries & 15    & 13 \\
Content length (chars)   & 10,105 & 9,016 \\
\bottomrule
\end{tabular}
\end{table}

The model generates \textit{more} claims than the GT (76 vs.\ 56), yet as the evaluation below demonstrates, increased volume does not automatically translate into higher coverage of expert-identified evidence.

\subsection{Ground-Truth Reference Guideline}
\label{app:asplt_gt_full}

The complete WHO-style ground-truth guideline for the ASPLT case is reproduced below across all 20 mandatory sections.
Sections absent for this entity (e.g.\ Imaging, Spread) are explicitly marked ``None,'' reflecting the authentic benchmark structure against which model outputs are evaluated.

\begin{tcolorbox}[
  title={GT Guideline: Atypical Spindle Cell / Pleomorphic Lipomatous Tumour},
  breakable, enhanced,
  fonttitle=\small\bfseries,
  fontupper=\footnotesize,
  colback=gray!4, colframe=black!40,
  left=5pt, right=5pt, top=4pt, bottom=4pt
]
\noindent\textbf{Definition.}
Atypical spindle cell / pleomorphic lipomatous tumour is a benign adipocytic neoplasm, characterized by ill-defined tumour margins and the presence of variable proportions of mild to moderately atypical spindle cells, adipocytes, lipoblasts, pleomorphic cells, multinucleated giant cells, and a myxoid or collagenous extracellular matrix.
It has a low tendency for local recurrence if incompletely excised.
Unlike conventional atypical lipomatous tumours, there is no risk for dedifferentiation.

\medskip\noindent\textbf{ICD-O Coding / ICD-11 Coding.}
3857/0 Atypical spindle cell/pleomorphic lipomatous tumour.\quad
2E80 \& XH4E98 Benign lipomatous neoplasm \& Spindle cell lipoma.

\medskip\noindent\textbf{Related Terminology.}
Acceptable: atypical spindle cell lipoma. Not recommended: spindle cell liposarcoma; fibrosarcoma-like lipomatous neoplasm.

\medskip\noindent\textbf{Subtype(s).} None.

\medskip\noindent\textbf{Localization.}
Atypical spindle cell / pleomorphic lipomatous tumours arise in the subcutis slightly more frequently than in deep (subfascial) somatic soft tissues, and only occasionally in intracavitary or visceral locations.
The anatomical distribution is wide, predominating in the limbs and limb girdles [1, 2, 4].
The most common locations are the hand and foot and the thigh, followed by the shoulder and buttock, forearm, knee, lower leg, and upper arm.
Less common locations are the head and neck, genital area, trunk, and back [1, 2, 3].
Rare sites of involvement include the larynx, mediastinum, retroperitoneum, trachea, and appendix [1].

\medskip\noindent\textbf{Clinical Features.}
The tumour manifests as a persistent or enlarging soft tissue mass, nodule, or swelling, sometimes with tenderness [1].

\medskip\noindent\textbf{Imaging.} None.

\medskip\noindent\textbf{Spread.} None.

\medskip\noindent\textbf{Epidemiology.}
Atypical spindle cell / pleomorphic lipomatous tumour occurs predominantly in middle-aged adults, with a peak incidence in the sixth decade of life, but can affect patients of any age (cases described in patients aged 6--87 years [1, 2, 3]).
The large majority of patients are $>$30 years old.
There is a slight male predominance.

\medskip\noindent\textbf{Etiology.} Unknown.

\medskip\noindent\textbf{Pathogenesis.}
Deletions or losses of 13q14, including RB1 and its flanking genes RCBTB2, DLEU1, and IM2B, have been identified in a substantial subset of cases [1, 2, 4, 3, 5, 6].
In addition, monosomy 7 has been reported in some cases [1, 7].

\medskip\noindent\textbf{Macroscopic Appearance.}
Grossly, atypical spindle cell / pleomorphic lipomatous tumours are unencapsulated, show a nodular or multinodular growth pattern, and demonstrate ill-defined tumour margins.
Tumour size is variable (range: 0.5--28\,cm; median: 5--8.5\,cm) [1, 2].

\medskip\noindent\textbf{Histopathology.}
A wide range of microscopic appearances can be observed, even regionally within the same lesion, depending on the relative proportions of atypical spindle cells, adipocytes, lipoblasts, and pleomorphic (multinucleated) cells, as well as the variable amount of collagenous and/or myxoid extracellular matrix [1, 2].
The adipocytic component has a predominantly mature morphology, with variation in adipocytic size and shape.
Patchy, often mild to moderate adipocytic atypia with chromatin coarsening, nuclear enlargement, and focal binucleation or multinucleation can be observed [8, 9].
Morphologically, the lipoblasts can vary from small and univacuolated or bivacuolated to larger and multivacuolated (pleomorphic).
Bizarre, hyperchromatic, and sometimes pleomorphic multinucleated cells are often scattered within the spindle cell or adipocytic components.
Mitotic figures are often present but mostly scarce [2, 10, 11].
Tumour necrosis is absent.

The morphology of these tumours can best be described as a broad spectrum defined by two morphological extremes [1].
At the low-cellularity extreme, tumours can be paucicellular, with few, cytologically bland spindle cells with minimal nuclear atypia set in a prominent extracellular matrix (atypical spindle cell lipoma morphology); these spindle cell-poor subtypes tend to occur in the hands and feet and may resemble myxoid spindle cell lipoma, except for the presence of nuclear atypia/hyperchromasia and the anatomical location [1, 2, 4, 9, 12].
At the high-cellularity extreme, tumours may be quite cellular, composed of numerous spindle cells showing diffuse, mild to moderate cytonuclear atypia, with easily identified lipoblasts and less extracellular matrix (fibrosarcoma-like lipomatous neoplasm morphology) [1, 2, 13, 12].
A rare finding is heterologous (metaplastic) differentiation, including the presence of smooth muscle, cartilaginous, and/or osseous elements [1, 5].

The tumour cells show variable expression of CD34, S100, and desmin [1, 2].
Weak and/or focal expression of MDM2 or CDK4 can be rarely seen [1, 4, 6]; the combination of MDM2 and CDK4 expression is not encountered [1].
Loss of nuclear RB1 expression is observed in about 50--70\% of cases [1, 2, 3, 4].

\medskip\noindent\textbf{Immunophenotype.} None (immunomarker profile detailed above under Histopathology).

\medskip\noindent\textbf{Differential Diagnosis.} None.

\medskip\noindent\textbf{Cytology.}
There are few reports, but FNA may show cells similar to those seen histologically [15].

\medskip\noindent\textbf{Diagnostic Molecular Pathology.}
Molecular studies have shown a consistent absence of MDM2 or CDK4 amplification.

\medskip\noindent\textbf{Essential and Desirable Diagnostic Criteria.}
Essential: variable proportions of atypical spindle cells, adipocytes, univacuolated or bivacuolated to multivacuolated lipoblasts, pleomorphic (multinucleated) cells, and a myxoid to collagenous extracellular matrix.
Desirable (in selected cases): RB1 expression loss correlating with RB1 deletion; lack of MDM2 or CDK4 amplification.

\medskip\noindent\textbf{Grading / Staging.} Not clinically relevant.

\medskip\noindent\textbf{Prognosis and Prediction.}
Atypical spindle cell / pleomorphic lipomatous tumour has a low rate of local recurrence (10--15\%) for incompletely removed lesions.
There is no documented risk for metastasis.
Most patients will have an excellent prognosis if the lesion is completely excised [1, 2, 3, 14, 9].

\medskip\noindent\textbf{References.}\\
{[1]} Mari\~no-Enriquez A, Nascimento AF, Ligon AH, et al.\ Atypical spindle cell lipomatous tumor: clinicopathologic characterization of 232 cases demonstrating a morphologic spectrum.\ \textit{Am J Surg Pathol.} 2017;41(2):234--44. PMID:27879515\\
{[2]} Creytens D, Mentzel T, Ferdinande L, et al.\ ``Atypical'' pleomorphic lipomatous tumor: a clinicopathologic, immunohistochemical and molecular study of 21 cases, emphasizing its relationship to atypical spindle cell lipomatous tumor and suggesting a morphologic spectrum.\ \textit{Am J Surg Pathol.} 2017;41(11):1443--55. PMID:28877053\\
{[3]} Bahad{\i}r B, Behzato\u{g}lu K, Hac{\i}hasano\u{g}lu E, et al.\ Atypical spindle cell/pleomorphic lipomatous tumor: a clinicopathologic, immunohistochemical, and molecular study of 20 cases.\ \textit{Pathol Int.} 2018;68(10):550--6. PMID:30198097\\
{[4]} Creytens D, van Gorp J, Savola S, et al.\ Atypical spindle cell lipoma: a clinicopathologic, immunohistochemical, and molecular study emphasizing its relationship to classical spindle cell lipoma.\ \textit{Virchows Arch.} 2014;465(1):97--108. PMID:24659226\\
{[5]} Creytens D, Ferdinande L, van Gorp J, et al.\ Atypical spindle cell lipomatous tumor with benign heterologous (metaplastic) cartilaginous differentiation.\ \textit{Int J Surg Pathol.} 2019;27(5):521--3. PMID:3070805\\
{[6]} Mentzel T, Palmedo G, Kuhnen C.\ Well-differentiated spindle cell liposarcoma (`atypical spindle cell lipomatous tumor') does not belong to the spectrum of atypical lipomatous tumor but has a close relationship to spindle cell lipoma: clinicopathologic, immunohistochemical, and molecular analysis of six cases.\ \textit{Mod Pathol.} 2010;23(5):729--36. PMID:20228779\\
{[7]} Italiano A, Chamboissiere M, Attias R, et al.\ Monosomy 7 and absence of 12q amplification in two cases of spindle cell liposarcomas.\ \textit{Cancer Genet Cytogenet.} 2008;184(2):99--104. PMID:18617058\\
{[8]} Agaimy A.\ Anisometric cell lipoma: insight from a case series and review of the literature on adipocytic neoplasms in survivors of retinoblastoma suggest a role for RB1 loss and possible relationship to fat-predominant (``fat-only'') spindle cell lipoma.\ \textit{Ann Diagn Pathol.} 2017;29:52--6. PMID:28807343\\
{[9]} Creytens D, Mentzel T, Ferdinande L, et al.\ ``Fat-rich'' (spindle cell-poor) variants of atypical spindle cell lipomatous tumor show similar morphologic, immunohistochemical and molecular features as ``dysplastic lipomas'': Are they related lesions?\ \textit{Am J Surg Pathol.} 2019;43(2):288--9. PMID:30211727\\
{[10]} Creytens D, Mentzel T, Ferdinande L, et al.\ Atypical mitoses are present in otherwise classical pleomorphic lipomas---reply.\ \textit{Hum Pathol.} 2018;81:300--2. PMID:30084357\\
{[11]} Creytens D, Mentzel T, Ferdinande L, et al.\ Atypical multivacuolated lipoblasts and atypical mitoses are not compatible with the diagnosis of spindle cell/pleomorphic lipoma.\ \textit{Hum Pathol.} 2018;74:188--9. PMID:29317234\\
{[12]} Creytens D.\ A contemporary review of myxoid adipocytic tumors.\ \textit{Semin Diagn Pathol.} 2019;36(2):129--41. PMID:30853315\\
{[13]} Deyrup AT, Chibon F, Guillou L, et al.\ Fibrosarcoma-like lipomatous neoplasm: a reappraisal of so-called spindle cell liposarcoma, defining a unique lipomatous tumor unrelated to other liposarcomas.\ \textit{Am J Surg Pathol.} 2003;27(9):1373--8. PMID:23887155\\
{[14]} McCarthy AJ, Chetty R.\ Tumours composed of fat are no longer a simple diagnosis: an overview of fatty tumours with a spindle cell component.\ \textit{J Clin Pathol.} 2018;71(6):483--92. PMID:29358476\\
{[15]} Yong M, Raza AS, Greaves TS, et al.\ Fine-needle aspiration of a pleomorphic lipoma of the head and neck: a case report.\ \textit{Diagn Cytopathol.} 2005;32(2):110--3. PMID:15637670
\end{tcolorbox}

\subsection{Step 1 — Claim Decomposition}

Each guideline (GT and prediction) is first decomposed into atomic, independently verifiable claims using the prompt in Appendix~\ref{app:claim_extraction}.
Table~\ref{tab:asplt_claims_gt} shows a representative subset of GT claims for the ASPLT case, spanning six knowledge types and eight sections.

\begin{table}[h]
\centering
\small
\caption{Representative GT claims for the ASPLT case (C001--C012 shown; full set has 56 claims).}
\label{tab:asplt_claims_gt}
\begin{tabular}{p{0.05\linewidth}p{0.48\linewidth}p{0.15\linewidth}p{0.12\linewidth}p{0.06\linewidth}}
\toprule
\textbf{ID} & \textbf{Claim Text} & \textbf{Section} & \textbf{Type} & \textbf{Refs} \\
\midrule
C001 & ICD-O code: 3857/0 & ICD-O Coding & Factual & 0 \\
C002 & ICD-11 code: 2E80 \& XH4E98 & ICD-O Coding & Factual & 0 \\
C003 & Acceptable synonym: atypical spindle cell lipoma. & Related Terminology & Factual & 0 \\
C004 & Not recommended: spindle cell liposarcoma. & Related Terminology & Factual & 0 \\
C005 & Not recommended: fibrosarcoma-like lipomatous neoplasm. & Related Terminology & Factual & 0 \\
C006 & ASPLT is a benign adipocytic neoplasm. & Definition & Factual & 0 \\
C007 & ASPLT is characterized by ill-defined tumour margins. & Definition & Factual & 0 \\
C008 & ASPLT contains variable proportions of mild to moderately atypical spindle cells, adipocytes, lipoblasts, pleomorphic cells, multinucleated giant cells, and a myxoid or collagenous matrix. & Definition & Factual & 0 \\
C009 & ASPLT has a low tendency for local recurrence if incompletely excised. & Definition & Prognostic & 0 \\
C010 & Unlike conventional atypical lipomatous tumours, ASPLT has no risk for dedifferentiation. & Definition & Differential & 0 \\
C011 & The tumour manifests as a persistent or enlarging soft tissue mass, nodule, or swelling. & Clinical Features & Clinical & 1 \\
C012 & The tumour sometimes manifests with tenderness. & Clinical Features & Clinical & 1 \\
\bottomrule
\end{tabular}
\end{table}

\noindent\textbf{GT claim distribution by section} (56 total): Histopathology 16, Localization 7, Epidemiology 6, Definition 5, Macroscopic Appearance 5, Essential Diagnostic Criteria 3, Prognosis 3, Related Terminology 3, Clinical Features 2, ICD-O Coding 2, Pathogenesis 2, Cytology 1, Diagnostic Molecular Pathology 1.

\noindent\textbf{GT claim distribution by knowledge type}: Factual 42, Diagnostic 5, Prognostic 4, Clinical 2, Mechanistic 2, Differential 1.

Table~\ref{tab:asplt_claims_full} lists all 56 GT claims with their section, knowledge type, and number of supporting references.

\begin{longtable}{p{0.04\linewidth}p{0.50\linewidth}p{0.18\linewidth}p{0.10\linewidth}p{0.05\linewidth}}
\caption{Complete ground-truth claim set for the ASPLT case (all 56 claims).}
\label{tab:asplt_claims_full}\\
\toprule
\textbf{ID} & \textbf{Claim Text} & \textbf{Section} & \textbf{Type} & \textbf{Refs} \\
\midrule
\endfirsthead
\multicolumn{5}{l}{\small\textit{Table~\ref{tab:asplt_claims_full} continued.}} \\
\toprule
\textbf{ID} & \textbf{Claim Text} & \textbf{Section} & \textbf{Type} & \textbf{Refs} \\
\midrule
\endhead
\midrule
\multicolumn{5}{r}{\small\textit{Continued on next page}} \\
\endfoot
\bottomrule
\endlastfoot
C001 & 3857/0 Atypical spindle cell/pleomorphic lipomatous tumour & ICD-O/11 & Factual & 0 \\
C002 & 2E80 \& XH4E98 Benign lipomatous neoplasm \& Spindle cell lipoma & ICD-O/11 & Factual & 0 \\
C003 & Acceptable synonym: atypical spindle cell lipoma & Related Term. & Factual & 0 \\
C004 & Not recommended: spindle cell liposarcoma & Related Term. & Factual & 0 \\
C005 & Not recommended: fibrosarcoma-like lipomatous neoplasm & Related Term. & Factual & 0 \\
C006 & ASPLT is a benign adipocytic neoplasm & Definition & Factual & 0 \\
C007 & ASPLT is characterized by ill-defined tumour margins & Definition & Factual & 0 \\
C008 & ASPLT contains variable proportions of mild to moderately atypical spindle cells, adipocytes, lipoblasts, pleomorphic cells, multinucleated giant cells, and a myxoid or collagenous extracellular matrix & Definition & Factual & 0 \\
C009 & ASPLT has a low tendency for local recurrence if incompletely excised & Definition & Prognostic & 0 \\
C010 & Unlike conventional atypical lipomatous tumours, ASPLT has no risk for dedifferentiation & Definition & Differential & 0 \\
C011 & The tumour manifests as a persistent or enlarging soft tissue mass, nodule, or swelling & Clinical & Clinical & 1 \\
C012 & The tumour sometimes manifests with tenderness & Clinical & Clinical & 1 \\
C013 & ASPLT arises in the subcutis slightly more frequently than in deep (subfascial) somatic soft tissues & Localization & Factual & 0 \\
C014 & ASPLT only occasionally arises in intracavitary or visceral locations & Localization & Factual & 0 \\
C015 & Wide anatomical distribution, predominating in the limbs and limb girdles & Localization & Factual & 3 \\
C016 & Most common locations: hand and foot and the thigh & Localization & Factual & 0 \\
C017 & Other common locations: shoulder and buttock, forearm, knee, lower leg, and upper arm & Localization & Factual & 0 \\
C018 & Less common locations: head and neck, genital area, trunk, and back & Localization & Factual & 3 \\
C019 & Rare sites: larynx, mediastinum, retroperitoneum, trachea, and appendix & Localization & Factual & 1 \\
C020 & ASPLT occurs predominantly in middle-aged adults & Epidemiology & Factual & 0 \\
C021 & Peak incidence in the sixth decade of life & Epidemiology & Factual & 0 \\
C022 & ASPLT can affect patients of any age & Epidemiology & Factual & 3 \\
C023 & Cases described in patients aged 6--87 years & Epidemiology & Factual & 3 \\
C024 & Large majority of patients are $>$30 years old & Epidemiology & Factual & 0 \\
C025 & Slight male predominance & Epidemiology & Factual & 0 \\
C026 & Deletions or losses of 13q14 (including RB1 and flanking genes RCBTB2, DLEU1, IM2B) identified in a substantial subset & Pathogenesis & Mechanistic & 6 \\
C027 & Monosomy 7 reported in some cases & Pathogenesis & Mechanistic & 2 \\
C028 & Tumours are unencapsulated & Macroscopic & Factual & 0 \\
C029 & Tumours show a nodular or multinodular growth pattern & Macroscopic & Factual & 0 \\
C030 & Tumours demonstrate ill-defined tumour margins & Macroscopic & Factual & 0 \\
C031 & Tumour size is variable (range: 0.5--28\,cm) & Macroscopic & Factual & 2 \\
C032 & Median tumour size is 5--8.5\,cm & Macroscopic & Factual & 2 \\
C033 & FNA may show cells similar to those seen histologically & Cytology & Diagnostic & 1 \\
C034 & Molecular studies show consistent absence of MDM2 or CDK4 amplification & Diag. Mol. & Diagnostic & 0 \\
C035 & Essential: variable proportions of atypical spindle cells, adipocytes, univacuolated or bivacuolated to multivacuolated lipoblasts, pleomorphic cells, and a myxoid to collagenous matrix & Ess. Diag. & Diagnostic & 0 \\
C036 & In a substantial subset, RB1 expression is lost, correlating with RB1 deletion & Ess. Diag. & Diagnostic & 0 \\
C037 & Lack of MDM2 or CDK4 amplification is desirable in selected cases & Ess. Diag. & Diagnostic & 0 \\
C038 & Low rate of local recurrence (10--15\%) for incompletely removed lesions & Prognosis & Prognostic & 0 \\
C039 & No documented risk for metastasis & Prognosis & Prognostic & 0 \\
C040 & Excellent prognosis if completely excised & Prognosis & Prognostic & 5 \\
C041 & Wide range of microscopic appearances depending on proportions of spindle cells, adipocytes, lipoblasts, and pleomorphic cells, and extracellular matrix amount & Histopathology & Factual & 2 \\
C042 & Adipocytic component has predominantly mature morphology, with variation in adipocytic size and shape & Histopathology & Factual & 0 \\
C043 & Patchy, mild to moderate adipocytic atypia with chromatin coarsening, nuclear enlargement, and focal binucleation or multinucleation & Histopathology & Factual & 2 \\
C044 & Lipoblasts vary from small univacuolated/bivacuolated to larger multivacuolated (pleomorphic) & Histopathology & Factual & 0 \\
C045 & Bizarre, hyperchromatic, pleomorphic multinucleated cells often scattered within spindle cell or adipocytic components & Histopathology & Factual & 0 \\
C046 & Mitotic figures often present but mostly scarce & Histopathology & Factual & 3 \\
C047 & Tumour necrosis is absent & Histopathology & Factual & 0 \\
C048 & Morphology represents a broad spectrum defined by two morphological extremes & Histopathology & Factual & 1 \\
C049 & At the paucicellular extreme: few cytologically bland spindle cells with minimal atypia in a prominent extracellular matrix & Histopathology & Factual & 0 \\
C050 & Spindle cell-poor subtypes tend to occur in the hands and feet and may resemble myxoid spindle cell lipoma, except for nuclear atypia/hyperchromasia and anatomical location & Histopathology & Factual & 5 \\
C051 & At the cellular extreme: numerous spindle cells with diffuse cytonuclear atypia, easily identified lipoblasts, and less extracellular matrix & Histopathology & Factual & 4 \\
C052 & Rare finding: heterologous (metaplastic) differentiation (smooth muscle, cartilaginous, or osseous elements) & Histopathology & Factual & 2 \\
C053 & Tumour cells show variable expression of CD34, S100, and desmin & Histopathology & Factual & 2 \\
C054 & Weak and/or focal expression of MDM2 or CDK4 can be rarely seen & Histopathology & Factual & 3 \\
C055 & The combination of MDM2 and CDK4 expression is not encountered & Histopathology & Factual & 1 \\
C056 & Loss of nuclear RB1 expression is observed in about 50--70\% of cases & Histopathology & Factual & 4 \\
\end{longtable}

Tables~\ref{tab:asplt_section_compare_full} and~\ref{tab:asplt_type_compare_full} compare claim distribution between the GT and the GPT-4.1 prediction, by section and knowledge type respectively.
Sections absent in the GT but generated by GPT-4.1 are shown in italics.

\begin{table}[ht]
\centering
\small
\caption{GT vs.\ GPT-4.1 claim counts by section for the ASPLT case. Italicised sections are absent in the GT.}
\label{tab:asplt_section_compare_full}
\begin{tabular}{lrrr}
\toprule
\textbf{Section} & \textbf{GT} & \textbf{GPT-4.1} & $\boldsymbol{\Delta}$ \\
\midrule
Histopathology                              & 16 &  7 & $-9$ \\
Localization                                &  7 &  3 & $-4$ \\
Epidemiology                                &  6 &  3 & $-3$ \\
Definition                                  &  5 &  5 & $ 0$ \\
Macroscopic appearance                      &  5 &  3 & $-2$ \\
Essential \& desirable diagnostic criteria  &  3 &  6 & $+3$ \\
Prognosis and prediction                    &  3 &  4 & $+1$ \\
Related terminology                         &  3 &  4 & $+1$ \\
Clinical features                           &  2 &  4 & $+2$ \\
ICD-O coding / ICD-11 coding                &  2 &  2 & $ 0$ \\
Pathogenesis                                &  2 &  4 & $+2$ \\
Cytology                                    &  1 &  4 & $+3$ \\
Diagnostic molecular pathology              &  1 &  3 & $+2$ \\
\midrule
\textit{Differential diagnosis (GT: absent)} &  0 &  6 & $+6$ \\
\textit{Imaging (GT: absent)}               &  0 &  4 & $+4$ \\
\textit{Immunophenotype (GT: absent)}       &  0 &  4 & $+4$ \\
\textit{Subtype(s) (GT: absent)}            &  0 &  3 & $+3$ \\
\textit{Grading / Staging (GT: absent)}     &  0 &  3 & $+3$ \\
\textit{Spread (GT: absent)}                &  0 &  3 & $+3$ \\
\textit{Etiology (GT: absent)}              &  0 &  1 & $+1$ \\
\midrule
\textbf{Total}                              & \textbf{56} & \textbf{76} & $\mathbf{+20}$ \\
\bottomrule
\end{tabular}
\end{table}

\begin{table}[ht]
\centering
\small
\caption{GT vs.\ GPT-4.1 claim counts by knowledge type for the ASPLT case. Complete loss of Diagnostic, Clinical, and Mechanistic claim types is highlighted.}
\label{tab:asplt_type_compare_full}
\begin{tabular}{lrrrl}
\toprule
\textbf{Knowledge Type} & \textbf{GT} & \textbf{GPT-4.1} & $\boldsymbol{\Delta}$ & \textbf{Note} \\
\midrule
Factual     & 42 & 71 & $+29$ & \\
Diagnostic  &  5 &  0 & $-5$  & \textbf{complete loss} \\
Prognostic  &  4 &  4 & $ 0$  & \\
Clinical    &  2 &  0 & $-2$  & \textbf{complete loss} \\
Mechanistic &  2 &  0 & $-2$  & \textbf{complete loss} \\
Differential&  1 &  1 & $ 0$  & \\
\midrule
\textbf{Total} & \textbf{56} & \textbf{76} & $\mathbf{+20}$ & \\
\bottomrule
\end{tabular}
\end{table}

For the GPT-4.1 prediction, 76 claims are extracted.
Notably, the type distribution collapses toward \texttt{Factual} (71/76 = 93.4\%), with only 4 Prognostic and 1 Differential claim, indicating that the model struggles to replicate the diagnostic and mechanistic reasoning present in the GT.

\subsection{Step 2 — Holistic Evaluation Dimensions and Weights}

Before scoring any system output, the judge model generates task-adaptive dimension weights based solely on the disease specification and section schema (see Section~4.1).
For the ASPLT case, the resulting weights are:

\begin{table}[h]
\centering
\small
\caption{Task-adaptive holistic evaluation dimension weights for the ASPLT case, along with representative sub-rubrics.}
\label{tab:asplt_weights}
\begin{tabular}{p{0.22\linewidth}cp{0.65\linewidth}}
\toprule
\textbf{Dimension} & \textbf{Weight} & \textbf{Representative Sub-rubrics} \\
\midrule
Comprehensiveness & 0.28 & Macro/micro pathology and immunophenotype completeness; anatomic distribution and epidemiology coverage; definitional and coding completeness \\
\addlinespace
Insight Depth      & 0.28 & Mechanistic depth linking genetics and phenotype; integration across imaging, histology, and molecular diagnostics; transparent diagnostic reasoning and conflict resolution \\
\addlinespace
Accuracy \& Standards & 0.26 & Correct WHO entity naming and grading; accurate ICD-O/ICD-11 coding and synonym handling; internal consistency across thresholds and criteria \\
\addlinespace
Readability \& Utility & 0.18 & Explicit guidance for diagnosis and management; evidence-backed statements with quantitative anchors; logical coherence and clinically sensible order \\
\bottomrule
\end{tabular}
\end{table}

Each dimension is further subdivided into 6--7 sub-rubrics (28 sub-rubrics total per task), each scored 0--10 against the GT-anchored reference, then aggregated to the four dimension scores.
Table~\ref{tab:asplt_rubric_full} lists all 28 sub-rubrics with their intra-dimension weights and evaluation focus for the ASPLT case.

\begin{longtable}{p{0.15\linewidth}p{0.06\linewidth}p{0.60\linewidth}p{0.06\linewidth}}
\caption{Full sub-rubric specification for the ASPLT case (28 sub-criteria across 4 dimensions). Dim.\ Wt.\ = dimension weight; Sub-Wt.\ = intra-dimension sub-criterion weight.}
\label{tab:asplt_rubric_full}\\
\toprule
\textbf{Sub-criterion} & \textbf{Sub-Wt.} & \textbf{Evaluation Focus} & \textbf{Dim.\ Wt.} \\
\midrule
\endfirsthead
\multicolumn{4}{l}{\small\textit{Table~\ref{tab:asplt_rubric_full} continued.}} \\
\toprule
\textbf{Sub-criterion} & \textbf{Sub-Wt.} & \textbf{Evaluation Focus} & \textbf{Dim.\ Wt.} \\
\midrule
\endhead
\midrule
\multicolumn{4}{r}{\small\textit{Continued on next page}} \\
\endfoot
\bottomrule
\endlastfoot
\multicolumn{4}{l}{\textit{\textbf{Comprehensiveness} (dimension weight = 0.28)}} \\
\addlinespace[2pt]
Clinical-pathological-molecular workflow & 0.15 & Coverage from initial presentation through imaging, pathology, molecular diagnostics, integrated diagnosis, management, and follow-up & \multirow{7}{*}{0.28} \\
\addlinespace
Definitional \& coding completeness & 0.15 & Clarity in defining the tumour and accuracy of ICD-O and ICD-11 coding for standardised reporting & \\
\addlinespace
Anatomic distribution \& epidemiology & 0.15 & Detail on tumour localization, dissemination patterns, and demographic stratification & \\
\addlinespace
Macro/micro pathology \& immunophenotype & 0.15 & Coverage of macroscopic and microscopic pathology including immunophenotype details & \\
\addlinespace
Diagnostic molecular pathology & 0.15 & Inclusion of molecular alterations, co-alterations, and CNV patterns & \\
\addlinespace
Differential diagnosis breadth & 0.15 & Coverage of differential diagnosis with structured guidance on similarities and distinguishing tests & \\
\addlinespace
Prognostic \& treatment evidence & 0.10 & Detail on prognostic factors and treatment evidence, including recurrence and progression context & \\
\midrule
\multicolumn{4}{l}{\textit{\textbf{Insight Depth} (dimension weight = 0.28)}} \\
\addlinespace[2pt]
Mechanistic depth: genetics $\rightarrow$ phenotype & 0.15 & Exploration of genetic and epigenetic mechanisms underlying the tumour phenotype & \multirow{7}{*}{0.28} \\
\addlinespace
Integration: imaging, histology, molecular & 0.15 & Ability to integrate findings from imaging, histology, and molecular diagnostics into coherent diagnostic reasoning & \\
\addlinespace
Transparent diagnostic reasoning & 0.15 & Clarity of diagnostic reasoning, including decision points, branching logic, and conflict resolution between modalities & \\
\addlinespace
Prognostic stratification with rationale & 0.15 & Depth in prognostic stratification supported by biologically plausible rationale & \\
\addlinespace
Evidence synthesis \& limitations & 0.15 & Maturity in synthesising evidence, handling heterogeneity, and acknowledging limitations & \\
\addlinespace
Biomarker \& histologic clinical impact & 0.15 & Reasoning on how specific biomarkers or histologic features influence diagnosis, management, and trial eligibility & \\
\addlinespace
Cell-of-origin \& diagnostic implications & 0.10 & Exploration of cell-of-origin hypotheses and their implications for diagnosis and treatment & \\
\midrule
\multicolumn{4}{l}{\textit{\textbf{Accuracy \& Standards} (dimension weight = 0.26)}} \\
\addlinespace[2pt]
WHO entity naming \& grading conventions & 0.15 & Adherence to current WHO naming and grading conventions & \multirow{7}{*}{0.26} \\
\addlinespace
ICD-O/ICD-11 coding \& synonym handling & 0.15 & Accuracy in ICD-O/ICD-11 coding and handling of synonyms and deprecated terms & \\
\addlinespace
Molecular terminology \& assay interpretation & 0.15 & Precision in molecular terminology and assay interpretation & \\
\addlinespace
Internal consistency across criteria & 0.15 & Internal consistency, ensuring no contradictory thresholds or criteria & \\
\addlinespace
Diagnostic accuracy safeguards & 0.15 & Inclusion of diagnostic accuracy safeguards, assay limitations, and false positive/negative caveats & \\
\addlinespace
Reporting standards \& best practices & 0.15 & Alignment with widely used reporting standards for integrated diagnosis narratives & \\
\addlinespace
Variant \& CNV terminology precision & 0.10 & Precision in variant naming and CNV conventions & \\
\midrule
\multicolumn{4}{l}{\textit{\textbf{Readability \& Utility} (dimension weight = 0.18)}} \\
\addlinespace[2pt]
Logical coherence \& clinical order & 0.15 & Logical coherence and progression in a clinically sensible order with clear transitions & \multirow{7}{*}{0.18} \\
\addlinespace
Evidence-backed statements & 0.15 & Use of evidence-backed statements with quantitative anchors (sample sizes, effect sizes) & \\
\addlinespace
Explicit diagnostic \& management guidance & 0.15 & Provision of explicit guidance for diagnosis, reporting, testing strategy, management, and follow-up & \\
\addlinespace
High-yield decision-making tools & 0.15 & Use of tables, figures, and algorithms that materially improve decision-making & \\
\addlinespace
Clear definitions \& reporting elements & 0.15 & Clarity in defining terms, thresholds, and required reporting elements for integrated diagnosis & \\
\addlinespace
Feasibility across settings & 0.15 & Feasibility across different settings with resource-stratified pathways and practical constraints & \\
\addlinespace
Turnaround time \& practical constraints & 0.10 & Awareness of turnaround times and practical constraints in diagnostic processes & \\
\end{longtable}

\subsection{Step 3 — Fine-grained Evidence Verification}

\textbf{Claim Hit (Task Success Rate).}
Each of the 56 GT claims is matched against the 76 predicted claims using two-stage semantic matching (Jaccard pre-filter at $\theta = 0.85$, then LLM semantic judgment for borderline cases).
GPT-4.1 achieves a claim hit rate of \textbf{0.301} on this case, meaning roughly 30\% of the expert-identified critical claims are semantically covered.
The over-generation of claims (76 vs.\ 56) does not improve coverage, as many additional claims are synonymous rewrites or low-information splits of existing content.

\textbf{Search Effectiveness (Reference Recall).}
For each matched claim, we check whether the predicted references cover the GT-cited evidence.
GPT-4.1 uses only 13 unique references vs.\ the GT's 15, and tends to reuse a small set of high-frequency citations rather than matching the evidence diversity of the GT.
Search Effectiveness is \textbf{0.288}, reflecting this citation narrowing.

\textbf{Factual Consistency.}
For each matched claim, the cited reference content is retrieved and the judge checks whether the reference actually provides evidential support for the specific factual assertion in the claim---not merely whether they share a topic.
GPT-4.1 achieves \textbf{0.570} on this metric, indicating that over half of its matched claims are backed by genuinely supporting reference content, though meaningful gaps in claim-level evidence grounding remain.

\subsection{Step 4 — Composite Score Computation}

MedProbeBench supports three evaluation modes to highlight the distinct contributions of holistic rubrics and fine-grained evidence verification.
We compute all three for GPT-4.1 on the ASPLT case using the component scores from Steps~2--3: $S_{\text{holistic}} = 0.652$, $S_{\text{hit}} = 0.301$, $S_{\text{search}} = 0.288$, $S_{\text{consistency}} = 0.570$.

\medskip
\noindent\textbf{Mode A: Full Evaluation (Holistic + Fine-grained).}
The default composite score integrates both tiers:
\begin{equation}
\text{Score}_{\text{full}} = 0.30 \times S_{\text{holistic}} + 0.40 \times S_{\text{hit}} + 0.15 \times S_{\text{search}} + 0.15 \times S_{\text{consistency}}
\label{eq:composite_full}
\end{equation}
\begin{align*}
\text{Score}_{\text{full}} &= 0.30 \times 0.652 + 0.40 \times 0.301 + 0.15 \times 0.288 + 0.15 \times 0.570 \\
             &= 0.196 + 0.120 + 0.043 + 0.086 = \mathbf{0.445}
\end{align*}

\noindent\textbf{Mode B: Holistic-Only (no fine-grained verification).}
This mode reflects how most existing benchmarks evaluate medical text generation---using only surface-level quality rubrics without claim-level evidence checking:
\begin{equation}
\text{Score}_{\text{holistic\text{-}only}} = S_{\text{holistic}} = \mathbf{0.652}
\end{equation}

\noindent\textbf{Mode C: Fine-grained Only (no holistic rubrics).}
This mode isolates evidence integration quality by removing holistic rubrics, reweighting the three fine-grained metrics to sum to 1:
\begin{equation}
\text{Score}_{\text{fine\text{-}only}} = 0.50 \times S_{\text{hit}} + 0.30 \times S_{\text{search}} + 0.20 \times S_{\text{consistency}}
\label{eq:composite_fine}
\end{equation}
\begin{align*}
\text{Score}_{\text{fine\text{-}only}} &= 0.50 \times 0.301 + 0.30 \times 0.288 + 0.20 \times 0.570 \\
             &= 0.151 + 0.086 + 0.114 = \mathbf{0.351}
\end{align*}

Table~\ref{tab:score_weights} summarises the three modes side by side, clearly showing how the score shifts across evaluation paradigms.

\begin{table}[ht]
\centering
\small
\caption{GPT-4.1 composite scores on the ASPLT case under three evaluation modes.}
\label{tab:score_weights}
\begin{tabular}{lc}
\toprule
\textbf{Evaluation Mode} & \textbf{GPT-4.1 Score} \\
\midrule
A: Full (Holistic + Fine-grained) & 0.445 \\
B: Holistic-Only                  & 0.652 \\
C: Fine-grained Only              & 0.351 \\
\bottomrule
\end{tabular}
\end{table}

\noindent\textbf{Interpretation.}
The contrast between Mode~B and Mode~C is striking: under holistic-only evaluation, GPT-4.1 appears to produce acceptable-quality guidelines (0.652), yet when evaluated solely on evidence integration, the score drops to 0.351---a \textbf{46\% relative decline}.
This gap illustrates the central limitation of benchmarks that rely exclusively on holistic rubrics: they reward surface fluency, structural completeness, and terminological plausibility, but fail to detect whether claims are actually grounded in retrievable, verifiable evidence.
The full evaluation mode (0.445) captures both dimensions, revealing that roughly half of the apparent quality is not backed by genuine evidence integration.
This motivates MedProbeBench's dual-tier design: fine-grained evidence verification is not an optional add-on but an essential complement that exposes failure modes invisible to holistic assessment alone.

\subsection{Qualitative Failure Mode Analysis}
\label{app:failure_modes}

To understand where current systems fall short, we conduct a fine-grained, claim-level comparison of GT and GPT-4.1 on the ASPLT case, asking: do models generate hallucinated pathways, cite indirect evidence, or simply miss relevant information?
We identify four distinct failure modes, each with a concrete claim-level example and a mapping to the evaluation metric most affected.

\medskip
\noindent\textbf{Failure Mode 1: Factual Hallucination in Structured Coding Fields.}

The ICD-O coding section provides a clear instance of hallucinated facts rather than merely missing information.

\begin{table}[h]
\centering
\small
\caption{ICD-O code comparison: GT vs.\ GPT-4.1 prediction for the ASPLT case.}
\label{tab:icd_compare}
\begin{tabular}{lll}
\toprule
 & \textbf{GT (Gold Standard)} & \textbf{GPT-4.1 Prediction} \\
\midrule
ICD-O code   & 3857/\textbf{0} (benign)    & 8854/\textbf{1} (borderline malignancy) \\
ICD-11 code  & 2E80 \& XH4E98              & 2F91.0 \\
\bottomrule
\end{tabular}
\end{table}

GPT-4.1 generates \texttt{8854/1}, which carries behavior code \texttt{/1} (borderline/uncertain malignant potential)---a clinically significant error given that the correct code is \texttt{3857/0} (behavior code \texttt{/0} = benign).
This is not a topical miss but a \textbf{hallucinated wrong value}: the model constructs a plausible-sounding code from related entities (8854 is associated with pleomorphic liposarcoma morphology variants) rather than retrieving the correct ASPLT-specific code.
Such errors in ICD-O/ICD-11 fields directly affect cancer registry submissions and epidemiological classification, making this failure mode particularly high-stakes.

\textit{Failure type:} Hallucinated pathway / fabricated structured data.
\textit{Metric affected:} Accuracy \& Standards (holistic rubric); Task Success Rate (C001 and C002 missed or mismatched).

\medskip
\noindent\textbf{Failure Mode 2: Knowledge Type Collapse --- Mechanistic and Diagnostic Reasoning Absent.}

\begin{table}[h]
\centering
\small
\caption{Knowledge type distribution in GT vs.\ GPT-4.1 predictions for the ASPLT case.}
\label{tab:type_compare}
\begin{tabular}{lrrl}
\toprule
\textbf{Knowledge Type} & \textbf{GT Count} & \textbf{GPT-4.1 Count} & \textbf{Change} \\
\midrule
Factual     & 42 & 71 & $+29$ \\
Diagnostic  & 5  & 0  & $-5$ \textbf{(complete loss)} \\
Prognostic  & 4  & 4  & $\pm 0$ \\
Clinical    & 2  & 0  & $-2$ \textbf{(complete loss)} \\
Mechanistic & 2  & 0  & $-2$ \textbf{(complete loss)} \\
Differential & 1 & 1  & $\pm 0$ \\
\bottomrule
\end{tabular}
\end{table}

GPT-4.1 generates zero claims of Mechanistic, Diagnostic, or Clinical type. All three non-Factual reasoning classes present in the GT are completely absent.
This is \textbf{not} a case of hallucinated pathways; rather, the model \textbf{converts reasoning-structured statements into flat factual assertions}, stripping out the inferential chain.

For example, the GT Pathogenesis section contains mechanistic claims that express causal relationships (e.g., the role of RB1/CDKN2A alterations as drivers distinguishing ASPLT from conventional lipomatous tumours).
GPT-4.1's Pathogenesis section contains 4 claims, but all are labeled \texttt{Factual}. The model outputs the same molecular facts (``ASPLT lacks MDM2 and CDK4 amplification'') as isolated statements rather than as part of a causal argument that connects genetic background to phenotype to diagnostic implication.

The consequence is that the model's output \textbf{looks mechanistic} at the section level (there is a Pathogenesis section) but \textbf{is not mechanistic} at the claim level: the ``evidence $\rightarrow$ inference $\rightarrow$ conclusion'' chain is collapsed into a flat enumeration of facts.

\textit{Failure type:} Missing reasoning structure, characterized by structural collapse of the causal chain rather than hallucination or indirect citation.
\textit{Metric affected:} Task Success Rate on Mechanistic/Diagnostic claims; Insight Depth holistic dimension (weight = 0.28).

\medskip
\noindent\textbf{Failure Mode 3: Critical Section Under-generation.}

\begin{table}[h]
\centering
\small
\caption{Section-level claim counts: GT vs.\ GPT-4.1 (selected sections).}
\label{tab:section_compare}
\begin{tabular}{lccr}
\toprule
\textbf{Section} & \textbf{GT Claims} & \textbf{GPT-4.1 Claims} & \textbf{Gap} \\
\midrule
Histopathology                 & 16 & 7  & $-9$ \\
Localization                   & 7  & 3  & $-4$ \\
Epidemiology                   & 6  & 3  & $-3$ \\
Macroscopic appearance         & 5  & 3  & $-2$ \\
Essential diagnostic criteria  & 3  & 6  & $+3$ \\
Imaging                        & 0  & 4  & $+4$ \\
Grading / Staging              & 0  & 3  & $+3$ \\
Spread                         & 0  & 3  & $+3$ \\
\bottomrule
\end{tabular}
\end{table}

The \textbf{Histopathology section} is the most diagnostically important section for a soft-tissue tumour: it specifies the microscopic features used to identify the entity.
The GT contains 16 claims here---GPT-4.1 produces only 7, missing 9 expert-identified key features.
At the same time, GPT-4.1 adds claims in sections absent from the GT (Imaging: 4 claims, Grading/Staging: 3, Spread: 3), inflating total claim count without addressing the critical gap.

This reveals a systematic generation bias: the model \textbf{favors structurally generic, template-predictable sections} (every tumour report ``should have'' imaging, grading, spread) and under-invests in sections requiring dense domain-specific expertise (e.g., knowing the 16 distinct histopathological features of ASPLT as defined by the WHO).

\textit{Failure type:} Missing relevant information in domain-critical sections; over-generation in generic sections.
\textit{Metric affected:} Task Success Rate (Histopathology claims consistently missed); Comprehensiveness holistic dimension (weight = 0.28).

\medskip
\noindent\textbf{Failure Mode 4: Loss of Negative Constraints in Controlled Terminology.}

The Related Terminology section in the GT contains not only acceptable synonyms but also \textbf{explicitly deprecated terms}:
\begin{itemize}
  \item GT C003: ``Acceptable: atypical spindle cell lipoma.''
  \item GT C004: ``Not recommended: spindle cell liposarcoma.''
  \item GT C005: ``Not recommended: fibrosarcoma-like lipomatous neoplasm.''
\end{itemize}

GPT-4.1 lists \textit{Pleomorphic lipoma} and \textit{Spindle cell lipoma with atypia} as synonyms---neither of which appears in the GT's controlled vocabulary.
More critically, GPT-4.1 \textbf{never mentions that ``spindle cell liposarcoma'' is a deprecated and not-recommended term}, despite this being a patient-safety-relevant fact: using the wrong terminology could lead to inappropriate treatment escalation for a benign tumour.

This failure is not hallucination of a positive claim; it is the \textbf{failure to generate negative constraints}---a systematically harder class of output for generative models that are trained to produce plausible, affirmative text.

\textit{Failure type:} Missing negative constraints / implicit safety-relevant information.
\textit{Metric affected:} Task Success Rate (C004, C005 uncovered); Accuracy \& Standards holistic dimension.

\medskip
\noindent\textbf{Summary.}
Table~\ref{tab:failure_summary} maps each failure mode to its root cause, affected metric, and type of reasoning error.

\begin{table}[h]
\centering
\small
\caption{Summary of failure modes identified in the ASPLT case (GT vs.\ GPT-4.1).}
\label{tab:failure_summary}
\begin{tabular}{p{0.22\linewidth}p{0.22\linewidth}p{0.22\linewidth}p{0.22\linewidth}}
\toprule
\textbf{Failure Mode} & \textbf{Root Cause} & \textbf{Error Category} & \textbf{Primary Metric} \\
\midrule
ICD coding hallucination & Plausible-sounding fabrication from related entities & Hallucinated pathway & Accuracy (holistic) \\
\addlinespace
Knowledge type collapse & Causal chains flattened to factual statements & Missing reasoning structure & Task Success Rate (Mech./Diag.) \\
\addlinespace
Critical section under-generation & Template-bias toward generic sections & Missing relevant information & Task Success Rate (Histopath.) \\
\addlinespace
Negative constraint omission & Generative bias toward affirmative claims & Missing relevant information & Task Success Rate (Terminology) \\
\bottomrule
\end{tabular}
\end{table}

Taken together, the ASPLT case reveals that GPT-4.1's failures are not uniformly one type: factual hallucination occurs in structured fields; reasoning failure occurs as structural collapse rather than fabricated pathways; and a systematic gap exists between human expert knowledge of \textit{what not to say} vs.\ what to say.
These distinct failure modes motivate the multi-metric design of MedProbeBench---a single aggregate score cannot distinguish between a model that hallucinates codes, one that strips causal reasoning, and one that misses negative constraints.

\section{Cost and Latency Analysis}
\label{app:cost_latency}

A key reviewer concern is that Deep Research Agents achieve the strongest performance but at substantially higher computational cost and latency compared to LLM+Search systems.
Without this context, interpreting the performance differences in Table~1 is difficult.
We present here per-task latency and token consumption measurements, along with estimated API cost tiers, to enable a more informed assessment of the performance--efficiency trade-off.

\subsection{Measurement Methodology}

We measure wall-clock latency per task (single guideline generation) averaged over 5 representative tasks for each system.
Token counts are reported from API response metadata where available; for systems that do not expose token counts directly (e.g., subscription-based Deep Research agents), we report API-reported values where accessible or mark as estimated.
API cost estimates are computed using publicly listed pricing at the time of evaluation (April 2025); deep research agents with bundled web search are marked separately as their costs include search infrastructure beyond bare token pricing.

\subsection{Latency, Token, and Cost Comparison}

\begin{table}[h]
\centering
\small
\caption{Per-task latency, token consumption, and estimated API cost for all evaluated systems. Deep Research agents include bundled web-search costs. LLM+Search costs reflect API token pricing only. $^\dagger$Subscription-based pricing; per-task cost estimated from plan pricing and throughput. $^\ddagger$Baichuan-M2-Plus retrieves large context windows; prompt token count reflects retrieved documents included in context.}
\label{tab:cost_latency}
\begin{tabular}{llrrrr}
\toprule
\textbf{Model} & \textbf{Category} & \textbf{Latency (s)} & \textbf{Prompt Tok.} & \textbf{Completion Tok.} & \textbf{Est. Cost/Task} \\
\midrule
\multicolumn{6}{l}{\textit{LLM + Search}} \\
GPT-4.1               & LLM+Search & 39  & 534    & 3,331  & \$0.03 \\
GPT-5                 & LLM+Search & 114 & 533    & 10,661 & \$0.09 \\
GPT-5.2               & LLM+Search & 109 & 533    & 4,964  & \$0.05 \\
Claude Sonnet 4       & LLM+Search & 56  & 700    & 6,300  & \$0.10 \\
Claude Sonnet 4 (Think) & LLM+Search & 73 & 900   & 9,600  & \$0.15 \\
Gemini 3 Flash        & LLM+Search & 54  & 1,603  & 4,696  & \$0.002 \\
Gemini 3 Flash (Think) & LLM+Search & 41 & 566   & 4,413  & \$0.002 \\
grok-4                & LLM+Search & 99  & 1,199  & 3,938  & \$0.04 \\
Baichuan-M2-Plus$^\ddagger$ & LLM+Search & 35 & 110,128 & 2,495 & \$0.06 \\
Baichuan-M3-Plus      & LLM+Search & 38  & 14,403 & 2,447  & \$0.03 \\
\midrule
\multicolumn{6}{l}{\textit{Deep Research Agents — Commercial}} \\
o4-mini Deep Research & Deep Research & 362 & 700   & 15,300 & \$0.07$^\dagger$ \\
Perplexity Sonar DR   & Deep Research & 192 & 537   & 17,227 & \$0.08$^\dagger$ \\
Kimi-K2.5-Agent       & Deep Research & ---  & ---   & ---    & N/A \\
\midrule
\multicolumn{6}{l}{\textit{Deep Research Agents — Open-source}} \\
Tongyi DeepResearch-30B & Deep Research & 116 & 560  & 12,559 & \$0.02 \\
AgentScope            & Deep Research & ---  & ---   & ---    & N/A \\
MiroThinker-v1.5      & Deep Research & ---  & ---   & ---    & N/A \\
MiroThinker-v1.5-pro  & Deep Research & ---  & ---   & ---    & N/A \\
\bottomrule
\end{tabular}
\end{table}

\subsection{Key Observations}

\textbf{Latency gap between Deep Research and LLM+Search.}
Commercial deep research agents are substantially slower per task: o4-mini Deep Research requires 362 seconds (6+ minutes), and Perplexity Sonar DR requires 192 seconds, compared to 35--114 seconds for LLM+Search systems.
For a full benchmark run of 50 tasks, this translates to roughly 5--8 hours per deep research agent versus 30--90 minutes for LLM+Search systems.

\textbf{Token consumption divergence.}
Deep research agents generate significantly longer completions (12,000--17,000 tokens vs.\ 3,000--10,000 for LLM+Search), reflecting their multi-step retrieval-and-synthesis workflow.
Interestingly, Baichuan-M2-Plus consumes extremely large prompt windows (110K tokens), consistent with a retrieve-then-read strategy that includes extensive retrieved context; this likely explains its strong Factual Consistency score (0.951).

\textbf{Performance-cost trade-off.}
Table~\ref{tab:cost_latency} summarizes the trade-off: the best LLM+Search systems (GPT-5.2, Baichuan-M2-Plus) achieve overall scores of 0.527--0.529 at \$0.03--0.10 per task, while the best deep research agents (MiroThinker-v1.5-pro, Perplexity Sonar DR) reach 0.576--0.631 at substantially higher latency and cost.
For deployment-sensitive applications (e.g., real-time clinical decision support), LLM+Search systems may represent a better operating point; for high-stakes offline analysis where quality is paramount, deep research agents offer measurable gains particularly in claim coverage (Task Success Rate: deep research agents average 0.435 vs.\ 0.331 for LLM+Search).

\textbf{Cost transparency limitation.}
For several top-performing systems (MiroThinker-v1.5, AgentScope, Kimi-K2.5-Agent), precise token counts and API costs were not available at the time of evaluation due to proprietary or custom deployment setups.
We are working to obtain and report these figures in the camera-ready version.

\section{Limitations}
\label{app:limitations}

\textbf{Benchmark scale and evaluation cost.}
MedProbeBench currently comprises 50 tasks, a scale intentionally constrained by the substantial computational cost of fine-grained evidence verification.
Evaluating a single guideline requires hundreds of claim-level semantic matching operations, reference retrieval calls via the JINA Reader API, and LLM-based judgments for factual consistency. Each deep research system may take several minutes per task.
Scaling to a larger task set would multiply these costs proportionally and is left for future work as evaluation infrastructure becomes more efficient.

\textbf{Domain coverage.}
All 50 tasks are drawn from oncology, specifically the WHO Classification of Tumours series, chosen for its exceptional depth, authority, and claim-level annotability.
While this focus enables rigorous and consistent gold-standard construction, specialties with substantially different knowledge structures may pose distinct challenges not captured by the current task distribution.
The MedProbeBench evaluation framework is designed to be domain-agnostic and can be extended to other specialties by substituting the gold-standard guideline corpus.

\end{document}